%% file: main.tex
\documentclass[10pt,twocolumn,letterpaper]{article}

\usepackage{wacv}              

\usepackage[dvipsnames]{xcolor}
\usepackage{graphicx}
\usepackage{amsmath}
\usepackage{amssymb}
\usepackage{booktabs}
\usepackage{marvosym}
\usepackage{orcidlink}
\usepackage{times}
\usepackage{epsfig}
\usepackage{graphicx}
\usepackage{amsmath}
\usepackage{amssymb}
\usepackage{booktabs}
\usepackage{multirow}
\usepackage{multicol}
\usepackage{algorithm}
\usepackage{algpseudocode}
\usepackage{pifont}%
\usepackage{enumitem}
\usepackage{wrapfig}
\usepackage{caption}
\usepackage{subcaption}
\newcommand{\cmark}{\color{ForestGreen}\ding{51}}%
\newcommand{\xmark}{\color{red}\ding{55}}%

\usepackage{adjustbox}
\usepackage{array}
\usepackage{makecell}
\usepackage{placeins}
\usepackage{listings}

\definecolor{codegreen}{rgb}{0,0.6,0}
\definecolor{codegray}{rgb}{0.5,0.5,0.5}
\definecolor{codepurple}{rgb}{0.58,0,0.82}
\definecolor{backcolour}{rgb}{0.95,0.95,0.92}

\lstdefinestyle{mystyle}{
    backgroundcolor=\color{backcolour},   
    commentstyle=\color{codegreen},
    keywordstyle=\color{magenta},
    numberstyle=\tiny\color{codegray},
    stringstyle=\color{codepurple},
    basicstyle=\ttfamily\footnotesize,
    breakatwhitespace=false,         
    breaklines=true,                 
    captionpos=b,                    
    keepspaces=true,                 
    numbers=left,                    
    numbersep=5pt,                  
    showspaces=false,                
    showstringspaces=false,
    showtabs=false,                  
    tabsize=2
}

\usepackage[capitalize]{cleveref}
\crefname{section}{Sec.}{Secs.}
\Crefname{section}{Section}{Sections}
\Crefname{table}{Table}{Tables}
\crefname{table}{Tab.}{Tabs.}

\def\wacvPaperID{649} %
\def\confName{WACV}
\def\confYear{2025}

\begin{document}

\title{ERM++: An Improved Baseline for Domain Generalization}

\author{Piotr Teterwak\footnotemark[3] \quad Kuniaki Saito\footnotemark[3] \quad Theodoros Tsiligkaridis\footnotemark[2] \quad Kate Saenko\footnotemark[3]  \quad Bryan A. Plummer\footnotemark[3] \\
 Boston University\footnotemark[3]  \qquad  MIT Lincoln Laboratory\footnotemark[2] \\
{\tt\small\{piotrt,keisaito,saenko,bplum\}@bu.edu}   
\quad{\tt\small ttsili@mit.edu} }
\maketitle

\begin{abstract}
Domain Generalization (DG) aims to develop classifiers that can generalize to new, unseen data distributions, a critical capability when collecting new domain-specific data is impractical. A common DG baseline minimizes the empirical risk on the source domains. Recent studies have shown that this approach, known as Empirical Risk Minimization (ERM), can outperform most more complex DG methods when properly tuned. However, these studies have primarily focused on a narrow set of hyperparameters, neglecting other factors that can enhance robustness and prevent overfitting and catastrophic forgetting, properties which are critical for strong DG performance. In our investigation of training data utilization (i.e., duration and setting validation splits), initialization, and additional regularizers, we find that tuning these previously overlooked factors significantly improves model generalization across diverse datasets without adding much complexity. We call this improved, yet simple baseline ERM++. Despite its ease of implementation, ERM++ improves DG performance by over 5\% compared to prior ERM baselines on a standard benchmark of 5 datasets with a ResNet-50 and over 15\% with a ViT-B/16. It also outperforms all state-of-the-art methods on DomainBed datasets with both architectures.  Importantly, ERM++ is easy to integrate into existing frameworks like DomainBed, making it a practical and powerful tool for researchers and practitioners. Overall, ERM++ challenges the need for more complex DG methods by providing a stronger, more reliable baseline that maintains simplicity and ease of use. Code is available at \url{https://github.com/piotr-teterwak/erm_plusplus}
\end{abstract}

\input{1_introduction}

\input{2_related_work}

\input{3_method}

\input{4_experiments}

\input{5_generalization}

\input{6_conclusion}

{\small
\bibliographystyle{ieee_fullname}
\bibliography{refs}
}
\clearpage
\input{10_arxiv_appendix}

\end{document}

%% file: 1_introduction.tex
\begin{figure}[t]
\centering
\includegraphics[width=0.85\linewidth]{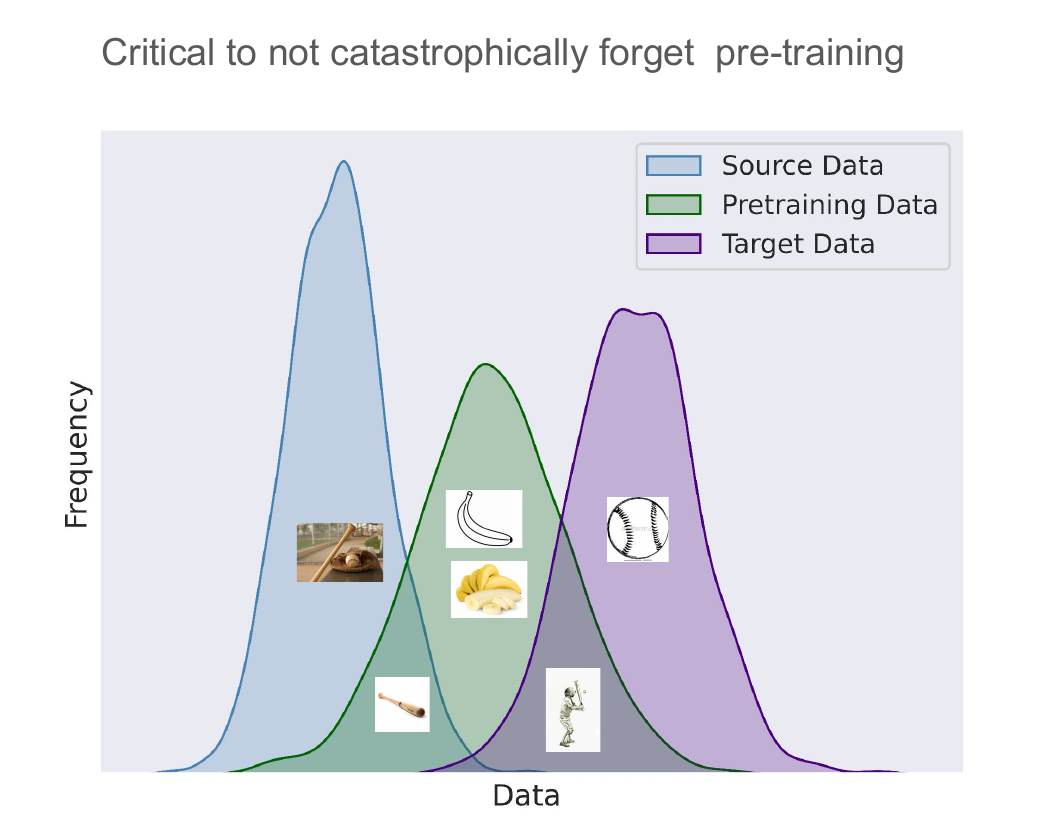}
\caption{Our goal is to provide a simple yet effective baseline that enhances robustness without adding complexity. Pre-training data (green) is often more similar to target data (purple) than source data (blue). Preventing catastrophic forgetting and overfitting to the source is thus critical in DG. We introduce ERM++, which addresses this through three key principles: Training Data Utilization (Sec. \ref{subsec:data_util}), Initialization (Sec. \ref{subsec:params}), and Regularization (Sec. \ref{subsec:weightspace_reg}).  }
\label{fig:fig1}
\vspace{-10pt}
\end{figure}
\section{Introduction}

Domain Generalization (DG) ~\cite{blanchard2011generalizing,muandet2013domain} tackles the challenge of developing models that can generalize to previously unseen data distributions without relying on the availability of new data distributions for model updates. This is vital when gathering new domain-specific data is impractical and differences between training and deployment data are unknown. In multi-source DG, each training sample is categorized as belonging to one of several domains. Many advanced methods extend Empirical Risk Minimization (ERM, ~\cite{vapnik1999overview}) by using this domain membership information ~\cite{ganin2016domain,zhang2021adaptive,li2018domain,zhou2021domain}.

\begin{figure}[t]
\centering
\includegraphics[width=0.75\linewidth]{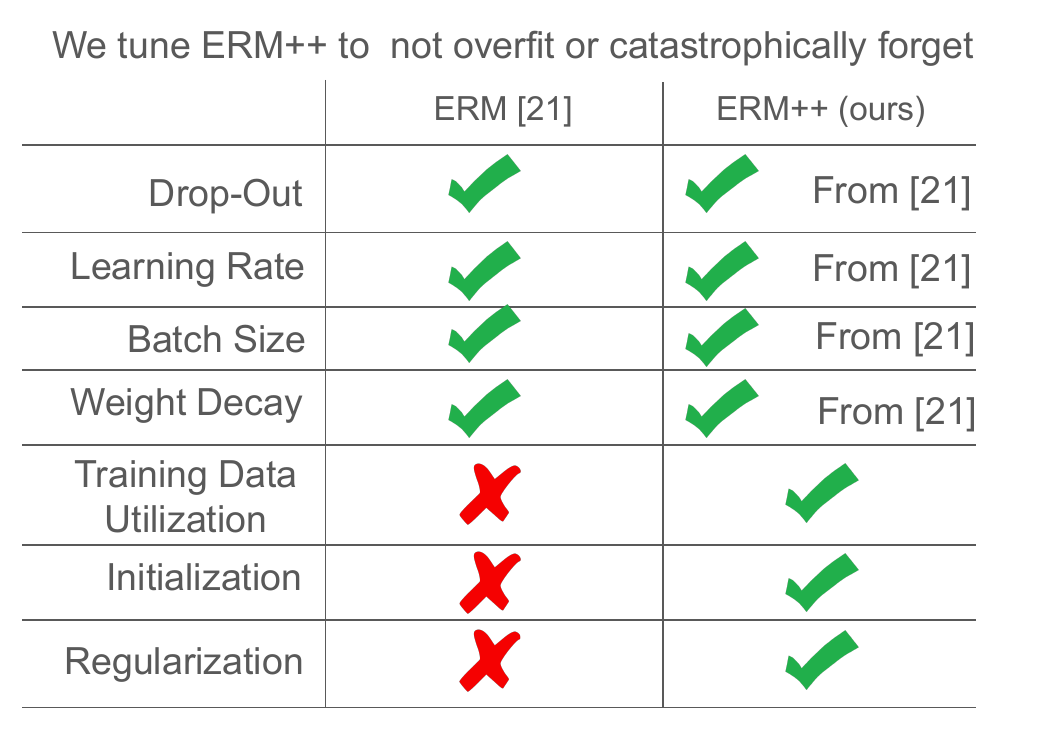}
\caption{Prior baselines like ERM \cite{gulrajani2020search} tune a small set of hyper-parameters. We extend tuning to other factors that control catastrophic forgetting and overfitting to the source. }
\vspace{-8pt}
\label{fig:fig2}
\end{figure}

However, recently, DomainBed~\cite{gulrajani2020search} conducted a comprehensive evaluation of these methods and revealed that ERM with hyper-parameter tuning outperformed the best numbers reported in the literature at the time. This shows the critical importance of well-tuned baselines; they ensure meaningful research results. Nevertheless, ERM~\cite{gulrajani2020search} tunes a small subset of important factors (Figure \ref{fig:fig2}) and overlooks others which prevent overfitting and catastrophic forgetting. 

Yet, the risk of overfitting and catastrophic forgetting remains a significant challenge in DG. DG methods divide training into two phases: pre-training and fine-tuning. Typically, DG methods take an off-the-shelf model pre-trained on some large-scale and generic dataset and fine-tune it on source data. Overfitting to source data is an especially large risk in DG; pre-training data distributions may be more similar to evaluation data than the available data for fine-tuning (See Figure \ref{fig:fig1}). For example, a DG model may be pre-trained on photographs, fine-tuned on sketches, and then evaluated on photographs again. Then, the model can easily forget pre-training data when fine-tuned on sketches, which is clearly detrimental to performance. Tuning settings that control overfitting to the source is the focus of our improved baseline, which we call ERM++. In extensive experiments, we show that this improves not only baseline performance over ERM\cite{gulrajani2020search} but also outperforms SOTA methods. 
We organize our additional tuning around three main principles: Training Data Utilization (Section \ref{subsec:data_util}), Initialization (Section \ref{subsec:params}), and Regularization (Section \ref{subsec:weightspace_reg}). 

\textbf{Training Data Utilization.} Modifying the number of samples seen during fine-tuning impacts overfitting. Currently, the DG standard is to cap the number of training steps for each method at a relatively low number while using a validation accuracy to select an exact checkpoint~\cite{gulrajani2020search,cha2021swad,cha2022domain}. While this can be effective in preventing strong overfitting, some methods may simply not converge. In addition, it can also limit the effectiveness of other regularizers by shortening training trajectories (see discussion of Model Parameter Average (\textit{MPA}) Section \ref{subsec:weightspace_reg}). Instead, we tune other hyper-parameters to prevent overfitting while automatically determining the training length and learning rate schedule customized for each model to help ensure convergence. We also use the training data more effectively than the DG standard. Increasing data size is known to improve generalization. Instead of splitting off a validation set and never using it to update the model, we do two passes. The first is to select the number of training steps with a validation set, and the second is to retrain for the selected number of steps with joined training and validation sets, increasing effective dataset size. 

\textbf{Initialization.} We consider how to initialize network parameters. Good initialization is known to impact convergence time \cite{NEURIPS2020_0607f4c7}, and strong pre-training procedures can minimize the distance from initialization to strong parameters for OOD domains. While it has been shown to be critical for domain adaptation and generalization~\cite{angarano2022back,kim2022broad}, initialization selection has not yet become part of a standard DG benchmark. We also investigate the role of pre-training data similarity to evaluation (target) data and find that pre-training data similarity is important for DG performance. Still, good initialization can help performance even for datasets very different from pre-training (Section \ref{subsec:beyond_web}).

\textbf{Regularization.} We expand prior works \cite{gulrajani2020search} exploration of regularization. Selecting which layers to tune and in which order is an important consideration in how far a fine-tuned network ends up from its initialization. A common example is first training the linear probe~\cite{kanavati2021partial,zhai2018classification}, but this has not been integrated into a commonly used multi-source DG baseline. Another form of weight-space regularization is averaging model iterates. This results in converging to flatter minima and improved generalization~\cite{arpit2021ensemble,cha2021swad,izmailov2018averaging,wortsman2022robust}. We investigate different averaging strategies in Section \ref{sec:results_weight_reg}.  

Put together, \textit{we find that the additional tuning in our improved baseline ERM++ outperforms both the prior ERM baselines and all recent SOTA methods on DomainBed}. ERM++ composes well with recent methods and has an attractively low compute cost compared to alternatives (Section \ref{subsec:cost}).  Applied to ViT-B models, a setting \cite{gulrajani2020search} did not explore, ERM++ outperforms ERM by 15\%. ERM++ even matches the performance of methods that leverage large-scale text encoders (CAR-FT\cite{mao2023context}) and update model parameters with test data (UniDG\cite{zhang2023unified}). Overall, our work underscores the importance of revisiting fundamental methods of controlling overfitting and challenges the need for more complex DG methods.

%% file: 2_related_work.tex
\vspace{-2pt}
\section{Related Work}

\noindent\textbf{Generalization in Neural Networks:} Generalization for neural networks has a long research history. Methods include ensembling \cite{fort2019deep,huang2016snapshot}, scaling of data\cite{kolesnikov2020big}, initialization for transfer learning\cite{kolesnikov2020big,oquab2024dinov,donahue2014decaf}, data augmentation\cite{cubuk2020randaugment}, and regularizing towards the initialization\cite{xuhong2018explicit}. Researchers have tried to correlate measures on the training with generalization to the test set\cite{jiang2019fantastic}. While many of these studies have been for in-domain generalization, their key ideas remain applicable for out-of-domain generalization. 
\smallskip

\noindent\textbf{Expanding Training  Data:}
Modifying the amount of training is a critical knob which controls generalization. Data augmentation is a common tool to expand the training data~\cite{zhou2021domain, hendrycks2020augmix, zhong2022adversarial,yan2020improve}. For example, 
Inter-domain mixup~\cite{yan2020improve} blends the images of different domains,  and augmentation with style transfer can further diversify training images~\cite{zhong2022adversarial}.  We  leverage a pre-trained model which already was trained with aggressive augmentation, AugMix \cite{hendrycks2020augmix}, but integrating stronger augmentation into ERM++ training is a promising direction for future work.

\noindent\textbf{Preventing catastrophic forgetting of good initializations with regularizers:} 
Several recent approaches aim to leverage generalizable features from a model pre-trained on large-scale data. Adapting such a model to the downstream task without forgetting its generalizable representations is the key to achieving generalization ~\cite{wortsman2022robust,cha2021swad,arpit2021ensemble} keep a running average of model parameters during training, which can be treated as an ensemble of the initialized model and fine-tuned model. ~\cite{kumar2022fine} and ~\cite{zhai2018classification} mitigate feature distortion by pre-training a linear probe first before fine-tuning the backbone, warmstarting the fine-tuning with a good initialization. MIRO~\cite{cha2022domain} maximizes the mutual information in feature space between the fine-tuned and pre-trained networks. We treat warm start and parameter averaging as hyper-parameters to add to ERM++ since their ease of implementation makes them suitable for a baseline.

\noindent\textbf{Domain-invariant feature learning:}
In multi-source domain generalization, it is common to leverage the domain labels to learn domain-invariant features. CORAL~\cite{sun2016deep} aligns second-order statistics of different domains. DANN~\cite{ganin2016domain} uses an adversarial loss to match feature distributions across source domains. 
However, these methods were evaluated with inconsistent baselines, and when ERM \cite{gulrajani2020search} accounted for this, advanced methods such as the above no longer showed a consistent advantage. We develop an even stronger baseline called ERM++.

%% file: 3_method.tex
\section{ERM++:  Improved ERM baseline for domain generalization}

We study the problem of Multi-Source DG for classification. 
We fine-tune a pre-trained model on data consisting of multiple domains and evaluate it on data from unseen domains. 
More formally, let us consider training domains $d\in \{d_1,...,d_n \}$. Each domain corresponds to a data source that is different from other domains; for example, domains may be "sketch" versus "real" images or different camera locations for wildlife cameras. A training dataset is constructed using all sample, label pairs in all training domains $D=\{(X^{d_1},Y^{d_1})...(X^{d_n},Y^{d_n})\}$.  After initializing classifier $f$ with a pretrained model, and finetuning on $D$, it is tested on a held-out testing domain $d_{test}$. 
We perform simple empirical risk minimization (ERM), formalized as minimizing the average loss over all samples $\frac{1}{n}\sum_{i\in D}\ell(x_i,y_i)$. We use mini-batch gradient descent and compose batches to be equal parts of each source domain. 

Our goal is to investigate the general training components that go into creating an ERM model to provide a strong baseline for future work, ensuring that improvements reported by new methodologies cannot be achieved using simpler means.  These include training data utilization (Section~\ref{subsec:data_util}), initialization (Section~\ref{subsec:params}), and regularization methods that help prevent overfitting to the source domains (Section~\ref{subsec:weightspace_reg}). We call our stronger baseline ERM++.

\subsection{Training Data Utilization: Setting Training Length Hyperparameters} 
\label{subsec:data_util}

A key component of training any neural network is utilizing the (often limited) training data effectively so that both underfitting and overfitting are avoided.  A common practice in the domain generalization literature is to split source datasets into (often 80\%/20\%) train/validation sets under a fixed number of iterations for each dataset (\eg,~\cite{gulrajani2020search,cha2021swad,rame2022diverse,arpit2021ensemble}). The validation data is used to set hyperparameters and perform checkpoint (no.\ training steps) selection to strike a balance between underfitting and overfitting.  This approach has two major drawbacks.  First, by creating a separate validation set, a significant portion of labeled data is sacrificed, and data quantity is known to be important for generalization.  Second, methods are trained under a fixed (relatively small) cap on number of iterations, resulting in underfitting. To address these issues, we use a two stage training pipeline, where we set training length hyperparameters in the first stage using a validation set, including training length and creating a learning rate schedule.  Then, in the second stage, we retrain the network using the entire dataset.  Additional details follow.

\noindent\textbf{Automatically determining training length \textit{(Auto-LR)}:} We observe that source validation performance does not saturate on many datasets using the settings in evaluation benchmarks used in prior work (See Supplementary), resulting in under-trained models.  We avoid this issue by automating the selection of training length as well as the learning rate schedule used on each dataset.    Specifically, we begin by splitting a dataset into its training and validation sets, which we use to set training length hyperparameters.  To determine the training length, we keep training a model until it does not improve performance on a validation set (held out from source domains) within the previous $N$ steps.  We then reduce the learning rate by 1/10th. Learning rates are reduced on the validation performance plateau twice, and we stop on the third drop (see supplementary for pseudocode). As this can provide variable training times, we do provide a high training step cap as an additional stopping condition (4x larger than DomainBed). Stopping after the third decrease helps avoid overfitting on some datasets while also allowing for longer training times and also provides a longer optimization trajectory for \textbf{Model Parameter Averaging} (See Section \ref{subsec:weightspace_reg}). 

\noindent\textbf{Using the full data \textit{(Full Data, FD)}:} After setting our training length and learning rate schedule in the first stage, stage two retrains the model using all our learned hyperparameters with the entire (train+validation) dataset.  This maximizes the amount of training data we use for a model to ensure we report its best performance with the available data.  In our experiments, we only set the training length and learning rate schedule once per model and reused those parameters for any other required tuning.  %

\subsection{Initializing Model Weights} 
\label{subsec:params}

 Domain generalization methods do not train a model from scratch, but rather transfer the weights of an existing model, \eg, ImageNet~\cite{deng2009imagenet} pretraining. However, prior work  assumes TorchVision model weights(\eg, ERM~\cite{gulrajani2020search}) for ResNet-50, which were trained using an outdated and basic training recipe. For ViT-B models, initializations beyond OpenAI CLIP\cite{radford2021learning} are rarely considered . Thus, another question we explore is what effect the data or training strategy has on DG performance.  We refer to the best initialization method from our analysis as \textbf{\textit{Strong Init}}, which represents the Augmix~\cite{hendrycks2020augmix} weights for ResNet-50 and DinoV2 \cite{oquab2024dinov} weights for ViT-B.

\subsection{Regularization}
\label{subsec:weightspace_reg}

Regularization has long been used to prevent the over-fitting of models to training data. Overfitting is especially problematic in DG since the source data has a different distribution than the target distribution. In fact, the pre-training data may be more similar to the target data than the source data. Thus, we explore several regularization techniques and focus on those which do not require an additional hyperparameter $\lambda$ in the loss.

\noindent\textbf{Model Parameter Averaging (\textit{MPA})} averages model iterates ~\cite{arpit2021ensemble,cha2021swad,izmailov2018averaging,ruppert1988efficient, wortsman2022model,wortsman2022robust,rame2022diverse,li2022branch}, which improves generalization by  converging to flatter minima~\cite{izmailov2018averaging}.  %
Arpit \etal~\cite{arpit2021ensemble} use a simple method for parameter averaging where simply all iterates are averaged (\textbf{\textit{MPA}}). Not only does \textit{MPA} converge to flatter minima, but it also averages model parameters from near the initialization and from near the converged minima, acting as regularizer by pulling the model closer to initialization. Furthermore, averaged parameters can viewed as an ensemble \cite{huang2016snapshot}. We verify that \textit{\textbf{MPA}} works in combination with other techniques present in ERM++. Finally,  long training enabled by \textbf{\textit{Auto-lr}} (another ERM++ component, Section \ref{subsec:data_util}) allows the optimization path to lengthen. This can improve the effectiveness of averaging model weights by increasing the diversity of averaged model parameters\cite{huang2016snapshot}. This is particularly intuitive with an ensemble view of weight averaging, as diverse ensembles are more performant than homogenous ones.  %

\noindent\textbf{Initializing Classifier Weights (\textit{Warm Start, WS}):} New class labels require a new classification layer, typically via random initialization. However, a recurring observation made by many researchers is that a model may suffer from divergence from the initialization due to the noisy, and large magnitude, gradients from  newly initialized layers~\cite{goyal1hour,he2016deep,rame2022diverse}. In the case of pretrained models, this results in catastrophic forgetting of robust, pre-trained features. To address this,  Warmstart is utilized (\textbf{\textit{WS}})~\cite{kanavati2021partial, zhai2018classification} (also commonly referred to as warmup), where the new layer weights are trained with all pretrained weights kept frozen for several hundred steps.   After this short training cycle, new and old layer weights are finetuned together (sometimes except for BatchNorm layers). This acts as a regularizer because after training the linear classifier, gradient norms are much smaller, resulting in parameters that are closer to the initialization after training. 

\noindent\textbf{Unfreezing BatchNorm for ResNets (\textit{UBN}):} Batch norm running statistics are frequently initialized from a pretrained model and kept frozen(\ie using batch normalization in evaluation mode)\cite{cha2021swad,gulrajani2020search,cha2022domain}. However, using training batch statistics (\ie using batch normalization in train mode) adds noise to the training, and acts as a regularizer. Therefore, we experiment with unfreezing batch norm. %

\noindent\textbf{Attention Tuning for ViTs (\textit{Attn Tune}):} Regularization methodologies such as Warm Start and Model Parameter Averaging are still applicable for ViT's, but Vision Transformers also offer additional options for weight regularization due to their unique architecture and properties. \cite{gandelsman2023interpreting} shows that the feed-forward features in Vision Transformers are hardly useful and that classification is done in the attention layers. Supporting this, \cite{touvron2022three} shows that attention-tuning is effective for standard transfer learning. By tying all weights that are not part of the attention layers explicitly to their initialized values, \textit{Attention Tuning} prevents many model parameters from being finetuned and, therefore, acts as a regularizer. We find it to be effective for ViTs.

%% file: 4_experiments.tex
\section{Experiments}
\label{sec:experiments}

We evaluate ERM++ using top-1 classification accuracy on a diverse set of datasets commonly used in multi-source DG, including \textbf{OfficeHome}~\cite{venkateswara2017deep}, \textbf{DomainNet}~\cite{peng2019moment}, \textbf{PACS}~\cite{li2017deeper},  \textbf{VLCS}~\cite{fang2013unbiased}, and \textbf{TerraIncognita}~\cite{beery2018recognition}; data and implementation are detailed in the Supplementary.  

\noindent\textbf{Implementation details.} We follow the DomainBed  training procedure, which is used for the ERM baseline,  and add additional components for ERM++. 
In particular, we use the default hyper-parameters from DomainBed~\cite{gulrajani2020search}, \eg, batch size of 32 (per-domain), learning rate of 5e-5, ResNet dropout value of 0, and weight decay of 0.  We train on all source domains except one, validate the model on held-out data from the sources every 300 steps, and evaluate on the held-out domain. The validation set does not contain any target data to avoid data leakage. 
\smallskip

\input{8_all_table}

\noindent\textbf{Results.} Table~\ref{table:all_methods} reports the performance of DG methods built on ERM across five DomainBed datasets.  Comparing the results using ERM++ on a ResNet backbone in Table~\ref{table:all_methods}(b) to methods from prior work in Table~\ref{table:all_methods}(a), we see ERM++ boosts performance by nearly 1\%.  Within Table~\ref{table:all_methods}(b) we see this boost can be increased by another 0.5\% by combining ERM++ with methods from prior work.  However, this gain is much smaller than methods like MIRO or DIWA combined with ERM (around a 4 point gain as seen in Table~\ref{table:all_methods}(a)).  We also note that MIRO and and DIWA are  more computationally expensive than ERM++ during training (Section \ref{subsec:cost}), and therefore ERM++ alone may be preferable in compute-constrained settings. 

Comparing the methods using a VIT-B backbone in Table~\ref{table:all_methods}(c), we see that ERM++ can match performance with OpenAI CLIP initialized methods that have an unfair advantage by either increasing their model size via a text encoder (CAR-FT), or use target data for model updates (UNI-DG).  Under a fairer comparison, ERM++ with a ViT-B backbone outperforms prior work by 5-17\%.  This demonstrates that by optimizing the training pipeline in ERM++, we can greatly boost performance, matching or exceeding the accuracy of more complicated, state-of-the-art methods.

 Below we analyze ERM++ components: training length (Section \ref{subsec:results_training_amount}), initialization (Section \ref{subsec:results_initialization}), and regularization (Section \ref{sec:results_weight_reg}). Section~\ref{subsec:cost} compares ERM++'s computational cost to prior work. Section~\ref{subsec:beyond_web} evaluates the effect of pretraining data and how well methods generalize to far out-of-domain settings using non-web-scraped data.

\input{7_table_erm++.tex}

\subsection{Training Data Utilization}
\label{subsec:results_training_amount}
\vspace{-2pt}
\noindent\textbf{Automatic Training Length/Learning Rate Schedule (\textit{Auto-LR}):} Section~\ref{subsec:data_util} describes our two stage training pipeline. In this first stage, we set the training length and learning rate schedule automatically using a validation set for each method.  This helps ensure we train until convergence, which can vary for each method and dataset. Training until convergence can have a large impact on transfer learning performance~\cite{chen2020simple}, and model averaging performs best when weights are diverse~\cite{rame2022diverse}, which benefits from the additional training time. At the same time, automatic stopping prevents over-fitting. In Table \ref{table:overall_table}(a) we show this boosts performance by 0.5 on average.%

\noindent\textbf{Using the full data (\textit{FD}):} The most common ERM implementation~\cite{gulrajani2020search} splits the training data into 80/20 train/validation splits, significantly reducing the size of the training set.  Thus, our second stage of training recombines the train/validation split and re-trains the model using the hyperparameters set in the first stage.
In Table \ref{table:overall_table}(a)  we show that this improves performance by up to 5\% on DomainNet. We emphasize that this is a fair comparison to prior work; validation data used for hyperparameter tuning (as in prior work) should also be considered training information.

\vspace{-2pt}
\subsection{Initializing Model Weights}
\label{subsec:results_initialization}

In Section~\ref{subsec:params} we note that prior work has often initialized their model weights using old models trained with out-of-date techniques.  In this section, we explore the effect of using \noindent\textbf{Stronger initializations (\textit{S.\ Init})} of model weights using more recent methods.  Table \ref{table:overall_table}(a) shows can obtain a 1\% boost on average, although we note that ERM++ still matches the performance of prior work reported in Table~\ref{table:all_methods}(a) when using the same initialization, but without requiring adding new loss functions requiring hyperparameter tuning (as in MIRO~\cite{cha2022domain}) or complicated training procedures (as with DIWA~\cite{rame2022diverse}).  However, selecting the best model takes some care.  To illustrate, let us consider the following initializations for ResNet-50.  \textbf{TorchVision Model Weights} is the standard ImageNet pretrained initialization present in TorchVision. It was trained with weak augmentations for 90 epochs. \textbf{AugMix~\cite{hendrycks2020augmix}}  is a method used to improve model consistency using augmentations without training the model on data which is too different from the test data. AugMix takes two augmented views of an image and mixes them in pixel space. Then, the model is trained to produce consistent output between two AugMix augmentations and the clean image. \textbf{ResNet A1}  initializes weights from the training recipe presented in ~\cite{wightman2021resnet}. The model is heavily tuned to find training settings that result in very strong ImageNet performance.  Examples include training for 600 epochs, the LAMB optimizer, strong augmentations, and a binary cross-entropy. \textbf{MealV2 ~\cite{shen2020mealv2}} is a highly performant ensemble, distilled into a ResNet-50. In particular, a SeNet-154 ~\cite{hu2018squeeze} (81.23\% ImageNet Top-1) and ResNet-152 (81.02\% ImageNet Top-1) are distilled into ResNet-50.

For ViT-B backbones, we consider the following. \textbf{ImageNet 1k\cite{steiner2022how}:} A recipe which is found to provide good results on ImageNet, making the model directly comparable to ImageNet pre-trained ResNet-50 model in terms of pre-training data similarity to Domain Generalization targets. \textbf{ImageNet-21k\cite{dosovitskiy2021an}}is the original ViT model. While the data scale is larger than ImageNet-1k,the pre-training domain is still real images. \textbf{OpenAI CLIP ViT-B/16\cite{radford2021learning}} is a strong ViT-B-16  model contrastively trained on 400 million image-caption pairs scraped from the web.  \textbf{DinoV2 ViT-B/14\cite{oquab2024dinov}} is a  self-supervised training method, trained on a highly curated dataset of 142 million samples.

Table \ref{table:inits} reports the performance of each initialization method, where we find that selecting a good checkpoint is not as simple as benchmarking ImageNet performance.
In particular, both the ResNet-50 A1 and Meal V2 weights achieve much better ImageNet Top-1 Accuracy than the standard TorchVision weights but do not achieve the best DG performance. The DG performance of the AugMix weights makes it a reasonable choice. %

\begin{table}[t]
\centering

\begin{tabular}{rl|c|c}
\small

&& Avg & ImageNet \\
\toprule
\textbf{(a)} & \textbf{ResNet-50 Backbone} & & \\
&TorchVision Weights & 66.6 & 76.1 \\
&AugMix~\cite{hendrycks2020augmix} & \textbf{68.4} & 79.0 \\
&Meal V2~\cite{shen2020mealv2} & 68.3 & \textbf{80.7} \\
&ResNet A1~\cite{wightman2021resnet} & 62.3 & 80.4 \\
\midrule
\textbf{(b)} & \textbf{ViT-B Backbone} & & \\
& ImageNet~\cite{dosovitskiy2021an} & 65.3 & 79.2 \\
& ImageNet-22k~\cite{steiner2022how}& 67.5 & \textbf{85.1} \\
& OpenAI CLIP~\cite{radford2021learning} & 76.9 & 80.2 \\
& %
DINOv2~\cite{oquab2024dinov} & \textbf{78.6} & 84.5 \\
\bottomrule
\end{tabular}
\caption{\textbf{DG performance with different ResNet-50 and ViT-B initialization.} \textbf{(a) ResNet-50:} The differences between initializations are very substantial. Interestingly, improved ImageNet accuracy does not strongly correlate with improved performance. %
\textbf{(b) ViT-B:} We find initialization is even more important here. ImageNet ViT models are weaker than ImageNet ResNet models, but scaling pre-training data (CLIP and DINO) improves performance by over 10\%.  }
\vspace{-13pt}
\label{table:inits}
\end{table}

We also observe that model distillation, which strongly improves source accuracy, does not increase overall DG performance. Meal V2 is a distillation of the ensemble of two very strong ImageNet models into a ResNet-50. Interestingly, the student in Meal V2 is initialized with the same AugMix-trained network as our experiments. Thus, the differences in performance can be strictly attributed to the effects of model distillation. Detailed results in the Supplementary show that ImageNet-like domains improve while performance on other domains suffers. This suggests that the distillation process effectively matches the student to the teacher over the data used in the distillation process, but not elsewhere.
AugMix is a model trained with generalization to synthetic corruptions as a goal, and it results in very strong DG performance. Thus, while ImageNet accuracy is not a good indicator of DG performance, investigating the correlation between synthetic corruption performance and DG performance is promising. 

From our transformer results in Table \ref{table:inits}(b), we generally find scaling pre-training data improves performance. OpenAI CLIP (400 million pre-training data points) outperform ImageNet-21k pretraining (14 million data points) by 10 percent, which improves over ImageNet pretraining (1.4 million data points) by 2 percent. However, training methodology can make up for dataset scale, as DinoV2 was only trained on 142 million data points yet achieves higher DG performance.  Still, the large pretraining datasets used by these models can be a large driver of performance by increasing similarity to DG performance(Section~\ref{subsec:beyond_web}).

\vspace{-3pt}
\subsection{Regularization}
\label{sec:results_weight_reg}

Table \ref{table:overall_table} reports the effect of the different regularization components we
discussed in Section~\ref{subsec:weightspace_reg}.  We find that each of the regularization components of model parameter averaging (MPA), warmstart (WS), unfreezing the batchnorm (UBN), and attention tuning (Attn Tune) all boosts performance by 0.5-3\%.  The largest gains come from MPA and Attn Tuning, which improves performance by 3\% and 1.5\%, respectively.  Most of the gain using Attn Tune comes from datasets with less available data, such as TerraIncognita's 5\% boost. We find that attention tuning also results in attention maps that attend to salient features, more so than full fine-tuning and especially more than just the pre-trained model (See Supplementary for examples).  We also note that by only finetuning the transformer's attention layers, we also reduce the memory required to train the network.
\smallskip

We also explored a setting where instead of averaging model weights, we attempt to include diversity between the models being averaged as this has been shown to boost performance~\cite{rame2022diverse}. Following ~\cite{li2022branch}, we first train a generalist model on all source domains for 28k steps, then train specialist models for 28k steps (1 model per domain), before averaging parameters. 
 Although averaging specialists improves over ERM by 2\%, it underperforms averaging iterates of a generalist by 2\%. More detailed results are reported in the supplementary. One possible explanation is that each domain has too little data.  %

\subsection{ERM++ Computational Cost} 
\label{subsec:cost}

\begin{table}[t]
\centering
\small

\setlength{\tabcolsep}{1pt}
\begin{tabular}{l|cc|c} \toprule
               & HP Searches                                                                & Train FLOPS   &  Top-1 \\
\midrule
ERM++ w/fewer steps & 2 (for \textbf{\textit{Auto-LR}})  & 1x            & 68.4\%           \\
ERM++         & 2 (for \textbf{\textit{Auto-LR}}) & \textless{}4x & \textbf{68.9}\%           \\
MIRO~\cite{cha2022domain}           & 4 (for $\lambda$)                                                    & 2x            & 68.1\%           \\
DIWA~\cite{rame2022diverse}           & 60                                                                               & 15x           & 68.0\% \\ \bottomrule         
\end{tabular}
\caption{Computational Cost: ERM++ achieves high performance without extensive hyper-parameter searches, instead using reasonable default ones. Even without an increased training step cap (Section \ref{subsec:data_util}), we're able to achieve SOTA performance. Train FLOPs are relative to ERM++w/out the increased training step cap of \textbf{\textit{Auto-LR}} Results are reported for ResNet-50. }
\label{table:computation}
\vspace{-10pt}
\end{table}

\input{11_image_similarity}

Table \ref{table:computation} compares the training cost of ERM++ to prior work.  We show that DIWA~\cite{rame2022diverse} and MIRO~\cite{cha2022domain} come at significant cost due to training a diverse ensemble or hyperparameter tuning for the contribution $\lambda$ of its new loss function, respectively.  In comparison, the only hyperparameters tuned for each are the training length and learning rate schedule in our two-stage process described in Section~\ref{subsec:data_util}.  This results in lower computational cost compared with prior work while also obtaining state-of-the-art results.

%% file: 8_all_table.tex
\begin{table*}[t!]
\small
\centering

\begin{tabular}{rlccccc|c} \toprule
\small
& & OfficeHome           & PACS                      & DomainNet                & Incognita                     & VLCS                      & {Avg.}         \\ \midrule
\textbf{(a)} & \multicolumn{5}{l}{\textbf{ ResNet-50 Backbone; built on ERM\cite{gulrajani2020search}}} &\\

& MLDG \cite{li2018learning}                                          & 66.8\scriptsize{$\pm0.6$} & 84.9\scriptsize{$\pm1.0$} & 41.2\scriptsize{$\pm0.1$} & 47.7\scriptsize{$\pm0.9$}          & 77.2\scriptsize{$\pm0.4$} & 63.6           \\
&ERM \cite{vapnik1999overview}                                      & 67.6\scriptsize{$\pm0.2$} & 84.2\scriptsize{$\pm0.1$} & 44.0\scriptsize{$\pm0.1$} & 47.8\scriptsize{$\pm0.6$}          & 77.3\scriptsize{$\pm0.1$} & 64.2           \\
&ERM+ MMD \cite{li2018mmd}                                                & 66.3\scriptsize{$\pm0.1$} & 84.7\scriptsize{$\pm0.5$} & 23.4\scriptsize{$\pm9.5$} & 42.2\scriptsize{$\pm1.6$}          & 77.5\scriptsize{$\pm0.9$} & 58.8           \\
&ERM + DANN \cite{ganin2016domain}                                         & 65.9\scriptsize{$\pm0.6$} & 83.6\scriptsize{$\pm0.4$} & 38.3\scriptsize{$\pm0.1$} & 46.7\scriptsize{$\pm0.5$}          & 78.6\scriptsize{$\pm0.4$} & 62.6           \\
&ERM + Mixup \cite{xu2020adversarial,yan2020improve} & 68.1\scriptsize{$\pm0.3$} & 84.6\scriptsize{$\pm0.6$} & 39.2\scriptsize{$\pm0.1$} & 47.9\scriptsize{$\pm0.8$}          & 77.4\scriptsize{$\pm0.6$} & 63.4           \\

&ERM + CORAL \cite{sun2016deep}                                            & 68.7\scriptsize{$\pm0.3$} & 86.2\scriptsize{$\pm0.3$} & 41.5\scriptsize{$\pm0.1$} & 47.6\scriptsize{$\pm1.0$}          & 78.8\scriptsize{$\pm0.6$} & 64.5           \\
\hline

&SagNet \cite{nam2021reducing}                                       & 68.1\scriptsize{$\pm0.1$} & 86.3\scriptsize{$\pm0.2$} & 40.3\scriptsize{$\pm0.1$} & 48.6\scriptsize{$\pm1.0$}          & 77.8\scriptsize{$\pm0.5$} & 64.2           \\
& mDSDI \cite{bui2021exploiting}                                                & 69.2\scriptsize{$\pm0.4$} & 86.2\scriptsize{$\pm0.2$} & 42.8\scriptsize{$\pm0.1$} & 48.1\scriptsize{$\pm1.4$}          & 79.0\scriptsize{$\pm0.3$} & 65.1           \\
&ERM + MIRO \cite{cha2022domain}                                                                        & 70.5\scriptsize{$\pm0.4$} & 85.4\scriptsize{$\pm0.4$} & 44.3\scriptsize{$\pm0.2$} & 50.4\scriptsize{$\pm1.1$}          & 79.0\scriptsize{$\pm0.0$} & 65.9           \\
&ERM + SWAD \cite{cha2021swad}                                                         & 70.6\scriptsize{$\pm0.2$} & 88.1\scriptsize{$\pm0.1$} & 46.5\scriptsize{$\pm0.1$} & 50.0\scriptsize{$\pm0.3$}          & 79.1\scriptsize{$\pm0.1$} & 66.9           \\
&ERM + CORAL + SWAD \cite{sun2016deep}                                                     & 71.3\scriptsize{$\pm0.1$} & 88.3\scriptsize{$\pm0.1$} & 46.8\scriptsize{$\pm0.0$} & 51.0\scriptsize{$\pm0.1$}          & 78.9\scriptsize{$\pm0.1$} & 67.3           \\
&ERM + DIWA \cite{rame2022diverse} & \textbf{72.8} & \textbf{89.0} &  \textbf{47.7} & 51.9 & 78.6 & 68.0  \\
&ERM + MIRO + SWAD  \cite{cha2022domain}                                                                  & 72.4\scriptsize{$\pm0.1$} & 88.4\scriptsize{$\pm0.1$} & 47.0\scriptsize{$\pm0.0$} & \textbf{52.9}\scriptsize{$\pm0.2$} & \textbf{79.6}\scriptsize{$\pm0.2$} & \textbf{68.1}           \\

\midrule
\textbf{(b)} & \multicolumn{5}{l}{\textbf{ResNet-50 Backbone; built on ERM++}} &\\
& ERM++ (w/out \textit{FD})                   & 74.0 &   89.6 & 45.8  & 51.0 &  78.3  & 67.7\\
& SWAD + ERM++ (w/out \textit{FD})                   & 74.1 &   89.6 & 49.7  & 49.7 &  77.3  & 68.0\\
& ERM++                                                                   & 74.7\scriptsize{$\pm0.0$}             & 89.8\scriptsize{$\pm0.3$}            & 50.8\scriptsize{$\pm0.0$}            & 51.2\scriptsize{$\pm0.3$}                              & 78.0\scriptsize{$\pm0.1$}                    & 68.9 \\

& DIWA + ERM++                   & 75.1 &   \textbf{90.0} &  \textbf{51.5} &  51.4   & \textbf{78.6} & 69.3\\
& MIRO + ERM++  &   \textbf{76.3}    &  88.8 &      50.4     &  \textbf{53.4}    &   77.9   & \textbf{69.4} \\
\midrule
\textbf{(c)} & \multicolumn{5}{l}{ \textbf{ViT-B  Backbone}} &\\
& ERM \cite{vapnik1999overview} & 66.4       & 83.4  & 44.4          & 35.3     & 75.9      & 61.1 \\
& ERM + MIRO\cite{cha2022domain} &   82.5     & 95.6  &    54.0     &   54.3   &   82.5    & 73.7 \\
& ERM + CAR-FT\cite{mao2023context}*        & 85.7  &    96.8     &   \textbf{62.5}   &  61.9     & 85.5 & 78.5 \\
& ERM + UNI-DG\cite{zhang2023unified}\Cross &   86.2    &  96.7 &   61.3      &  62.4   &  \textbf{86.3}    &  \textbf{78.6}\\
& ERM++                  & \textbf{86.7} & \textbf{96.8} & 59.8 & \textbf{67.4}    &    82.4  & \textbf{78.6} \\
\bottomrule
\end{tabular}
\vspace{-5pt}
\caption{\textbf{ERM++ Domain Generalization Performance:} \textbf{(a)} We show 
DG performance of methods with an ERM foundation. \cite{gulrajani2020search} provided an ERM baseline, on which new methods like MIRO and SWAD could be built. All methods below the horizontal line were developed after the tuned ERM baseline\cite{gulrajani2020search} was published, showing how it pushed the DG field forward. ERM++  provides a refreshed baseline for future work to do the same.   \textbf{(b)} Performance of our ERM++ baseline. We show that ERM++ is an even stronger foundation for recent methods than ERM. SWAD is built on top of ERM++ w/out FD because it requires validation data for interval selection. \textbf{(c)} When we extend ERM++ to vision transformers, ERM++ outperforms ERM by a full 15\%, showing the generality of ERM++ principles. *CAR-FT\cite{mao2023context} utilizes the text encoder of CLIP for regularization, which adds 63 million parameters to the model and is not present in all ViT-B models. \Cross UNI-DG utilizes target data for model updates, so it falls outside of the bounds of DG,  but we add it for reference.  }
\label{table:all_methods}
\vspace{-10pt}
\end{table*}

%% file: 7_table_erm++.tex
\begin{table*}[t]
\centering
\small

\begin{tabular}{cllccccc|c}
\toprule
&Backbone/Category& Ablation & OfficeHome & PACS & DomainNet & TerraIncognita & VLCS & {Avg.} \\ \midrule
\midrule
\textbf{(a)} & \textbf{ResNet-50 Backbone}& ERM++ & \textbf{74.7} & 89.8  & \textbf{50.8}  & 51.2 & 78.0 & \textbf{68.9} \\ 
\midrule
&Train Length& w/o \textit{Auto-LR} & 74.6 & 87.9  & 49.8 &  51.1 & 78.6 & 68.4 \\ 
&Train Length& w/o \textit{FD} & 74.0 & 89.6 & 45.8 &  51.0 & 78.3 & 67.7 \\ 
\midrule
&Initialization& w/o \textit{S. Init} & 72.6 & 88.8 & 48.6 & 49.2  & 78.7  & 67.6 \\ 
\midrule
&Regularization& w/o \textit{UBN} & 74.7 & \textbf{90.1} & 49.9 &  49.0 & 78.6 & 68.3 \\ 
&Regularization& w/o \textit{WS} & 74.3 &89.5 & 50.7 & \textbf{51.9} & 77.2 & 68.7 \\
&Regularization& w/o \textit{MPA} & 71.5 & 72.3 & 46.9 & 48.9 & \textbf{79.8} & 65.9\\
\midrule
\midrule
\textbf{(b)} & \textbf{ViT B/16 Backbone}& ERM++ & \textbf{86.7} & \textbf{96.8} & 59.8 & \textbf{67.4}    &    \textbf{82.4}  & \textbf{78.6} \\
\midrule
&Regularization& w/o \textit{Attn Tune} & 85.6 &  95.6 & \textbf{60.3} &  62.6   &  81.1    & 77.0 \\
 \bottomrule
\end{tabular}

\caption{We present the overall ablation for ERM++. \textbf{(a)} Each component contributes to the strong performance of ERM++. with  \textit{Model Parameter Averaging} being the most impactful part of ERM++. The use of \textit{Full Data}. \textit{Model Parameter Averaging}, and \textit{Strong Init} are most critical, while the other methods give small boosts which add up (see cumulative study in the supplementary). \textbf{(b)} We also ablate the ViT-specific component of \textit{Attention Tuning}. We find that \textit{Attention Tuning} improves performance on almost datasets, except for DomainNet, which has the largest amount of data. The jump is especially large on TerraIncognita, of about 5\%. }
\label{table:overall_table}
\vspace{-10pt}
\end{table*}

%% file: 11_image_similarity.tex
\begin{table*}[t]
 \small
 \centering

\begin{tabular}{rl|ccccc|cc}
\hline
             &                           & OfficeHome    & PACS & DomainNet & TerraIncognita & VLCS & FMOW & PCAM \\ \hline
\textbf{(a)} & Image Similarity          & 79.8    & 77.3 & 73.6      & 66.1      & 69.9 & 70.4 & 76.9 \\
             & Text Similarity           & 83.1      & 85.8   & 83.2        & 83.0        & 86.4   & 77.2   & 71.3 \\
             & Average Similarity        & 81.4    & 81.7 & 78.3      & 74.6      & 78.0 & 73.7 & 74.0 \\
             & Approximate Duplicates    & 8\%     & 9\%  & 7.5\%     & 0\%       & 0\%  & 0\%  & 0\%  \\ \hline
\textbf{(b)} & ERM++ w/ResNet-50 Init.   & 74.7    & 89.8 & 50.8      & 52.9      & 79.6 & 48.1 & 89.4 \\
             & ERM++ w/DINOv2 Init.     & \textbf{86.7}    & \textbf{96.8} & 59.8      & \textbf{67.4}      & 82.4 & \textbf{54.8} & \textbf{95.9} \\
             & ERM++ w/OpenAI CLIP Init. & 85.9    & 96.2 & \textbf{61.1}      & 59.5      & \textbf{83.2} & 48.2 & 93.7 \\ 
            & OpenAI CLIP zeroshot      & 82.3    & 96.1 & 57.7      & 23.1      & 82.4 & 19.7 & 55.3 \\ \hline
\end{tabular}
 \caption{\textbf{Relationship between dataset similarity and DG performance.} We compute the similarity of LAION-400M for DomainBed (left) and Stanford-WILDS (right) datasets.  Overall, performance gains over ImageNet pretraining are largest on datasets where  text  similarity is high to LAION-400M, suggesting pre-training data plays a key role in DG performance. Nevertheless, DINOv2 init performs strongly on dissimilar data as well.  More discussion in Section \ref{subsec:beyond_web}. }
\label{table:similarities}
\vspace{-14pt}
\end{table*}

%% file: 5_generalization.tex
\vspace{-5pt}

\section{Generalizing Beyond Web-scraped Datasets} 
\label{subsec:beyond_web}

We have demonstrated that ERM++ is a highly effective recipe for DG on a suite of common DomainBed benchmarks. However, OfficeHome, PACS, and DomainNet are all web-scraped datasets. Because pre-training of foundation ViTs is also on very large scale web-scraped data, it is important to understand how much of their improved performance is driven by train-test similarity. Therefore, we expand our evaluation to include two out-of-domain (OOD) datasets: FMOW~\cite{koh2021wilds,christie2018functional} and PCAM~\cite{koh2021wilds,bandi2018detection}.  We also measure similarity of   LAION-400M~\cite{schuhmann2021laion}(a representative pre-training dataset) to  DG datasets to provide additional insight. We find that dataset similarity is a strong driver of performance, but that the strong DINOv2 initialization can deliver high performance even on OOD datasets.

\noindent\textbf{Metrics.} In addition to hold-one-out DG performance on models with several different initializations, we also measure similarity between target datasets and a representative large-scale web-scraped pre-training datasets, LAION-400M. To do so, we use an approximate nearest nearest neighbors algorithm \cite{beaumont-2022-clip-retrieval}, and build a retrieval index with OpenAI CLIP-B/32 image embeddings. We then measure dataset similarity by querying the index with dataset samples from the DG dataset, and record the cosine similarity of the nearest neighbor. Averaged across 1000 random samples, this constitutes an \textit{image similarity}. We  do the same with  class labels of the target datasets, to create a \textit{text similarity}. We combine the two modalities by averaging, creating a \textit{average similarity}. We also probe the DG for exact duplicates within LAION-400M, by sampling a random 200 images from each downstream dataset,and retrieving their nearest neigbor using the image index described above. We manually inspect these images for exact duplicates (except for slight crop differences) between query and retrieval. 

\noindent\textbf{Results.} Table~\ref{table:similarities}(a) reports the similarity between pretraining dataset LAION-400M~\cite{schuhmann2021laion} and the images in the DG benchmark datasets.  We observe a number of interesting findings. First, 3 of 4 datasets (DomainNet, OfficeHome, PACS) with the highest average similarity with LAION-data also contain the largest number of duplicates. Second, the average similarities score of DomainBed datasets is much higher than that of the Stanford-WILDS datasets, suggesting that most current benchmarks evaluate Domain Generalization results that are not very different from large-scale pre-training. Third, in Table~\ref{table:similarities}(b) we find that even without fine-tuning with ERM++,  CLIP zero-shot performance is very high on all the datasets which have high average similarity. This suggests CLIP is a very strong initialization even without any finetuning. However, ERM++ fine-tuning is critical for all datasets which have poor zero-shot performance. Finally, ERM++ with CLIP and DINOv2 initializations improve most over ImageNet initializations (ResNet-50) on datasets which have high \textit{text similarity} to LAION-400M. The exception is VLCS, which consists of all real domains, and therefore is also similar to ImageNet, in which case ImageNet pre-training may be sufficiently strong. This suggests that as long as concepts are present in the pre-training, ERM++ fine-tuning with foundational ViTs can overcome image dissimilarity between fine-tuning and pre-training images.  Put together, these observations indicate that pre-training data similarity, especially in text-space, to target domains is a big driver of ERM++ performance. Nevertheless, the \textbf{\textit{Strong Init}} (DINOv2) for ViT-B boosts performance over OpenAI clip init on the dissimlar WILDS-FMoW by 7\% and WILDS-PCAM by 2\%, suggesting that pre-training methodology complements pre-training data composition, and delivers strong performance.
\vspace{-1pt}

%% file: 6_conclusion.tex
\vspace{-16pt}
\section{Conclusion}

This paper develops a strong yet simple baseline, ERM++, for improving the performance of DG models. By identifying several techniques for enhancing ERM, our approach achieves significant gains in DG performance, reporting a 5\% boost over ERM and 1\% better than the SOTA on ResNet-50. We find ERM++ improves over ERM even more significantly when applied to unstable and over-parameterized ViT models, by over 15\%. Our results highlight the importance of improving the training procedure for enhanced DG performance. 

\noindent\textbf{Future Extensions of ERM++:} While ERM++ demonstrates significant improvements, it also opens up several promising avenues for further exploration. For instance, integrating stronger data augmentation techniques could be a natural extension to improve generalization. Additionally, we leverage hyper-parameters from \cite{gulrajani2020search}, but searching for better values may further improve performance.

\noindent\textbf{Acknowledgements:} DISTRIBUTION STATEMENT A. Approved for public release. Distribution is unlimited.

This material is based upon work supported by the Under Secretary of Defense for Research and Engineering under Air Force Contract No. FA8702-15-D-0001. Any opinions, findings, conclusions or recommendations expressed in this material are those of the author(s) and do not necessarily reflect the views of the Under Secretary of Defense for Research and Engineering.

%% file: 10_arxiv_appendix.tex
\section{Additional Results}

\subsection{Cumulative Study}

In addition to the ablative study of ERM++ components in the main paper, we add a cumulative study in Table \ref{table:cumulative_table}. Similar to the ablative study, we can see that performance is enhanced by each component of ERM++, in some cases by over 2\% (Experiment 2, MPA). 

\begin{table*}[t]
\begin{center}
\scriptsize

\begin{tabular}{c|c|c|c|c|c|c|ccccc|c}
\toprule
\multicolumn{7}{c|}{\small  Cumul. study (\#6 is full ERM++) }                   & \multirow{2}{*}{OfficeHome }     & \multirow{2}{*}{PACS}           & \multirow{2}{*}{VLCS}           & \multirow{2}{*}{DomNet}      & \multirow{2}{*}{TerraInc} & \multirow{2}{*}{Avg.}       \\     \cmidrule{1-7} 

\#&MPA&FD&WS&Auto-LR&S. Init& UBN&15K&10K &11K  &590K &25K \\ \midrule
1 & \xmark & \xmark  & \xmark & \xmark & \xmark & \cmark & 67.1\scriptsize${\pm0.2}$ & 85.1\scriptsize${\pm0.3}$
          & 76.9\scriptsize${\pm0.6}$ & 44.1\scriptsize${\pm0.15}$ & 45.2\scriptsize${\pm0.6}$ & 63.7 \\ 
2 & \cmark &  \xmark & \xmark & \xmark & \xmark & \cmark & 70.2\scriptsize${\pm0.3}$
          & 85.7\scriptsize${\pm0.2}$          & 78.5\scriptsize${\pm0.3}$
& 46.4\scriptsize${\pm0.0}$
          & 49.4\scriptsize${\pm0.4}$
          & 66.0 \\ 
3 & \cmark  & \cmark & \xmark & \xmark & \xmark & \cmark          & 71.5\scriptsize${\pm0.1}$
 & 87.3\scriptsize${\pm0.2}$ & 77.4\scriptsize${\pm0.1}$
          & 46.8\scriptsize${\pm0.0}$ & 49.8\scriptsize${\pm0.5}$          & 66.5          \\ 
4 & \cmark  & \cmark & \cmark & \xmark & \xmark &  \cmark & 72.6\scriptsize${\pm0.1}$      & 88.8\scriptsize${\pm0.1}$ & 77.0\scriptsize${\pm0.1}$                                                & 48.6\scriptsize${\pm0.0}$     & 49.3\scriptsize${\pm0.3}$          & 67.3    \\

 5 & \cmark  & \cmark & \cmark & \cmark & \xmark & \cmark  & 72.6\scriptsize${\pm0.1}$ & 88.8\scriptsize${\pm0.1}$ & \textbf{78.7\scriptsize${\pm0.0}$} & 48.6\scriptsize${\pm0.0}$ & 49.2\scriptsize${\pm0.3}$          & 67.6 \\ 
 6 & \cmark  & \cmark & \cmark & \cmark & \cmark & \cmark & \textbf{74.7\scriptsize${\pm0.0}$}            & \textbf{89.8}\scriptsize${\pm0.3}$            & 78.0\scriptsize${\pm0.1}$            & \textbf{50.8\scriptsize${\pm0.0}$}                              &  \textbf{51.2\scriptsize${\pm0.3}$}                    & \textbf{68.9} \\ 
 \bottomrule
\end{tabular}
\end{center}
\caption{We present the overall cumulative study for ERM++. ERM++ corresponds to experiment 6. (1) ERM~\cite{gulrajani2020search} baseline with unfrozen BN. (2) MPA: Model parameter averaging, which uniformly improves results. 
(3) FD: training on the full data.  
(4) WS: Warm-starting the classification layer especially improves OfficeHome and PACS. (5) Auto-LR: Learning the lr schedule, which ensures convergence improves performance by an additional half percent.
 (6) S.Init: Initializing the initial parameters to those trained with AugMix brings performance to state of the art. }  \label{table:cumulative_table}
\end{table*}

\subsection{MealV2 Distillation Results}

In Table \ref{table:distill}, we look at the per-domain accuracy on DomainNet, comparing Augmix training (Aug) and MealV2 (MV2). MealV2 is a method used to distill a large ensemble into a student ResNet-50, where the student is initialized to AugMix weights.  We can see that the distillation process, while dramatically improving ImageNet performance, only slightly changes DG performance. In particular, generalization gets slightly worse for all domains except for (R)eal, which is the most similar to ImageNet. This is surprising, since it has been shown that both ensembles ~\cite{arpit2021ensemble} and larger models ~\cite{angarano2022back} improve DG performance.

\begin{table*}[t]
\centering
\begin{tabular}{l|cccccc|c}
\hline
        & Painting       & Clipart       & Info         & Real          & Quickdraw     & Sketch         & Avg            \\ \toprule
Aug~\cite{hendrycks2020augmix}  & \textbf{57.3} & \textbf{68.8} & \textbf{25.6} & 70.2          & \textbf{17.1} & \textbf{59.8} & \textbf{49.8} \\ 
MV2~\cite{shen2020mealv2} & \textbf{57.3} & 68.5          & 25.4          & \textbf{70.9} & 16.1          & 59.0          & 49.5          \\ \bottomrule
\end{tabular}
\caption{\textbf{Model distillation's effect on DG: }We look at the per-domain accuracy on DomainNet, comparing Augmix training (Aug) and MealV2 (MV2). MealV2 is a method used to distill a large ensemble into a student ResNet-50, where the student is initialized to AugMix weights.  We can see that the distillation process, while dramatically improving ImageNet performance, only slightly changes DG performance.}
\label{table:distill}
\end{table*}

\subsection{Weight space regularization}
In Table \ref{table:weight_reg} we explore a setting where instead of averaging model weights, we attempt to include diversity between the models being averaged as this has been shown to boost performance~\cite{rame2022diverse}. Following ~\cite{li2022branch}, we first train a generalist model on all source domains for 28k steps, then train specialist models for 28k steps (1 model per domain), before averaging parameters. 
 Although averaging specialists improves over ERM by 2\%, it underperforms averaging iterates of a generalist by 2\%. One possible explanation is that each domain has too little data to create a good specialist model

\begin{table*}[t]
  \centering

  \begin{tabular}{l|cccccc|c} \toprule
    & Painting & Infograph & Quickdraw & Sketch & Real & Clipart & Avg \\ \midrule
    ERM & 51.1 & 21.2 & 13.9 & 52.0 & 63.7 & 63.0 & 44.1 \\
    SMPA & 52.9 & \textbf{27.2} & 14.3 & 51.3 & 65.6 & 65.2 & 46.1 \\
    MPA & \textbf{55.2} & 24.0 & \textbf{16.7} & \textbf{57.4} & \textbf{67.0} & \textbf{67.49} & \textbf{48.0} \\ \bottomrule
  \end{tabular}
  \caption{\textbf{Weight Space Regularization:} We experiment with different types of parameter averaging for weight regularization on DomainNet. \textbf{SMPA} is a specialized model parameter averaging, where we average parameters of domain specialists, while \textbf{MPA} averages parameters within a single training trajectory. While both outperform ERM, \textbf{MPA} outperforms \textbf{SMPA}. %
}
  \label{table:weight_reg}
\end{table*}

\subsection{Per-dataset details}

In Tables \ref{tab:oh} (OfficeHome), \ref{tab:dn} (DomainNet), \ref{tab:vlcs} (VLCS), \ref{tab:ti} (TerraIncognita), \ref{tab:pacs} (PACS), we expand results for the datasets  and report accuracies for each held-out domain.  We compare ResNet-50 ERM++ with reported performances of ERM \cite{gulrajani2020search}, DIWA \cite{rame2022diverse},  SWAD,  \cite{cha2021swad}, and MIRO ~\cite{cha2022domain}. ERM + SWAD + MIRO  and DIWA are the current SOTA for ResNet-50 models for this set of datasets. Overall trends include ERM++ being especially effective at sketch-like domains, indicating a lowered texture bias. On the sketch and clipart domains in DomainNet, ERM++ outperforms prior best performance by over 4\%. When we additionally combine MIRO with ERM++, we see much improved performance on OfficeHome and TerraIncognita without much affecting the performance on the other datasets.

\input{9_tables_supp_no_se.tex}

\subsection{Validation-Set Accuracy Curves}
\label{subsec:val_curves}

\begin{figure*}[t]
\includegraphics[width=\linewidth]{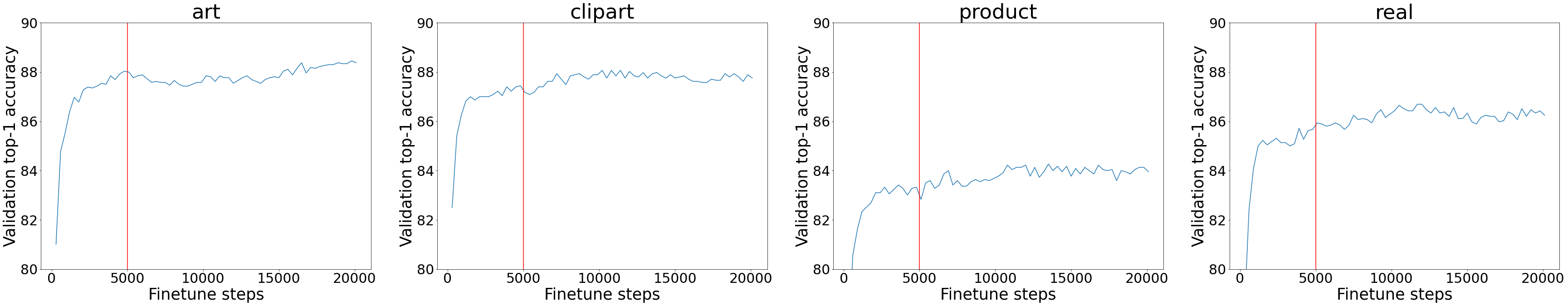}
\caption{\textbf{OfficeHome:} Source validation accuracies. The validation accuracy saturates by 20000 steps. Training length used in prior works is denoted as a red line, and the training is not yet converged. }
\label{fig:oh_val}
\end{figure*}

\begin{figure*}[t]
\includegraphics[width=\linewidth]{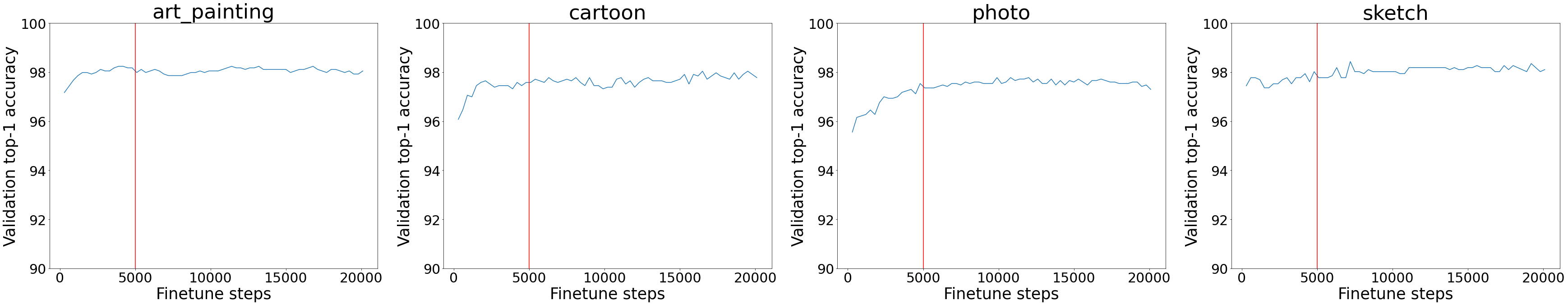}
\caption{\textbf{PACS}: Source validation accuracies. The validation accuracy saturates by 20000 steps. Training length used in prior works is denoted as a red line, and the training is not yet converged.  }
\label{fig:pacs_val}
\end{figure*}

\begin{figure*}[t]
\includegraphics[width=\linewidth]{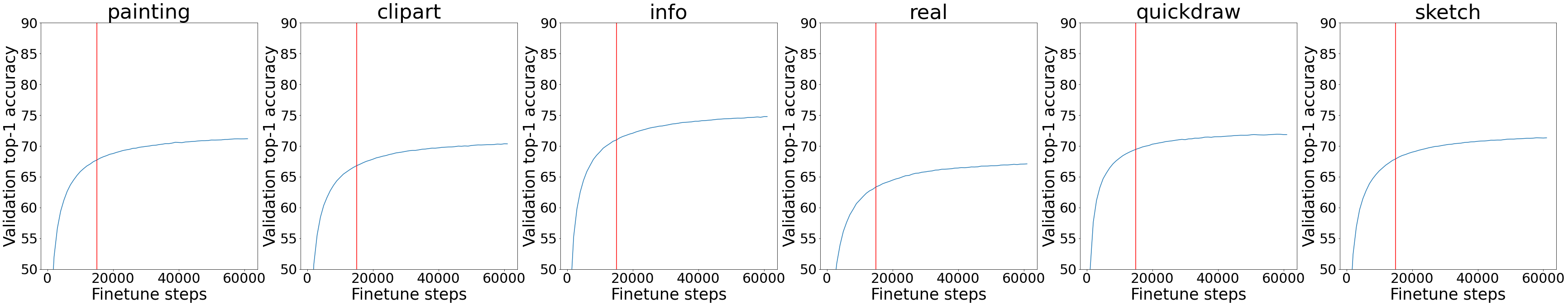}
\caption{\textbf{DomainNet:} Source validation accuracies. The validation accuracy saturates by 60000 steps. Training length used in prior works is denoted as a red line, and the training is not yet converged. }
\label{fig:dn_val}
\end{figure*}

\begin{figure*}[t]
\includegraphics[width=\linewidth]{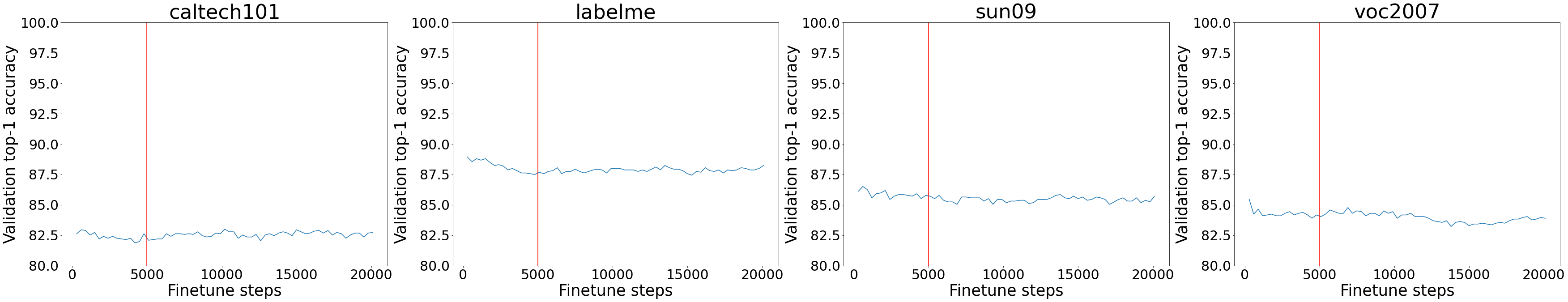}
\caption{\textbf{VLCS:} Source validation accuracies. The validation accuracy saturates by 20000 steps. Training length used in prior works is denoted as a red line. In the case of VLCS, it seems like longer training is not so helpful, and this shows the need for \textbf{\textit{Auto-lr}}}. 
\label{fig:vlcs_val}
\end{figure*}

\begin{figure*}[t]
\includegraphics[width=\linewidth]{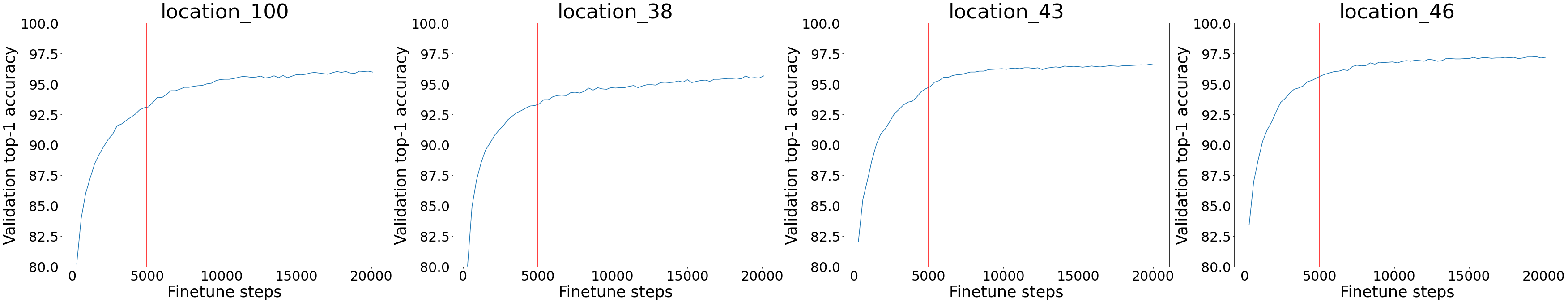}
\caption{\textbf{TerraIncognita:} Source validation accuracies. The validation accuracy saturates by 20000 steps. Training length used in prior works is denoted as a red line, and the training is not yet converged. }
\label{fig:ti_val}
\end{figure*}

In Figures \ref{fig:oh_val},\ref{fig:pacs_val},\ref{fig:dn_val},\ref{fig:vlcs_val}, and  \ref{fig:ti_val}, we provide source-validation accuracies for each of the 5 datasets, for 20000 steps for most datasets except for the larger DomainNet, which is 60000 steps. As one can see, at this point, validation accuracy is saturated for most domains in most datasets, so this training length is reasonable. Prior training lengths are denoted as red vertical lines in these figures, and one can see that for many datasets this is not a sufficient training length.  As we describe in Section 4.1 of the main paper, extending training  lengths with \textbf{\textit{Auto-LR}} improves performance by  0.5\% on average.

\begin{figure*}[h!]
    \begin{lstlisting}[language=Python]
import torch
import torch.optim as optim
import torch.nn as nn

# Initialize variables for learning rate decay
lr_decay_factor = 0.1
lr = initial_lr
max_decay_steps = 3
decay_count = 0

# Training loop
for step in range(num_steps):
    # Your training code here
    
    # Validation
    if (step %
        model.eval()
        validation_loss = compute_validation_loss(model, validation_data, criterion)
    
        # Check if validation loss is not decreasing
        if validation_loss >= previous_validation_loss:
            decay_count += 1
            if decay_count >= max_decay_steps:
                break
            else:
                # Decay learning rate
                lr *= lr_decay_factor
                for param_group in optimizer.param_groups:
                    param_group['lr'] = lr
    
        # Update previous validation loss for the next iteration
            previous_validation_loss = validation_loss
    \end{lstlisting}
    \caption{\textbf{Learning Rate Decay with \textbf\textit{{Auto-lr}}}. The learning rate is decreased by a factor of 0.1 when the validation loss does not decrease. Training is stopped after the third consecutive non-decreasing validation loss.}
    \label{fig:learning_rate_decay}
\end{figure*}

\section{Dataset Visualizations}

In Figures \ref{fig:oh_data} (OfficeHome),
\ref{fig:dn_data} (DomainNet), \ref{fig:vlcs_data} (VLCS), 
\ref{fig:ti_data} (TerraIncognita),
\ref{fig:pacs_data} (PACS),   \ref{fig:fmow_data} (FMoW), and \ref{fig:pcam_data} (PCAM) we show samples of a few classes from each of the datasets, and each domain. As one can see, both the datasets and distribution shifts are quite diverse, highlighting the flexibility of our method. We present some key attributes of the datasets below. 
\smallbreak
\noindent\textbf{OfficeHome~\cite{venkateswara2017deep}} Figure \ref{fig:oh_data}. This dataset focuses on household objects.  The domain shifts are in low-level style mostly, and there is little spatial bias.
\smallbreak
\noindent\textbf{DomainNet~\cite{peng2019moment}} Figure \ref{fig:dn_data}.   While the real domain is quite similar to what one might expect in ImageNet, the distribution shifts are quite substantial in other domains. Quickdraw and Infograph are particularly challenging, so the 1-3\% gains of ERM++ on these domains is meaningful (Table \ref{tab:dn}).

\noindent\textbf{VLCS~\cite{fang2013unbiased}:} Figure \ref{fig:vlcs_data}. Low-level statistics are quite similar between domains in this dataset, however spatial biases differ between domains. For example, Caltetch objects are quite centered, while other domains do not have this trait. For example the LabelMe domain has cars along the side of the image, and there are many chairs in the VOC2007 domain. Furthermore, in some cases the size of the objects differs dramatically.  Lastly, there are many ambiguous images in the LabelMe domain (see Figure \ref{fig:broken_vlcs}), raising questions about the validity of trying to improve performance on this dataset.

\smallbreak
\noindent\textbf{TerraIncognita~\cite{beery2018recognition}}: Figure \ref{fig:ti_data} The background stays consistent, and the animal object frequently takes up a small portion of the frame. At night the images are black-and-white.  This is a very realistic dataset, on which is good to test.
\smallbreak
\noindent\textbf{PACS~\cite{li2017deeper}} Figure \ref{fig:pacs_data}. The subjects tend to be centered, and the sketches are more realistic than the quickdraw setting in DomainNet. Though the domains are similar to that of DomainNet, PACS has fewer than 10000 samples compared to 586000 of DomainNet. Therefore PACS tests the capabilities of ERM++ on smaller data.

\smallbreak
\noindent\textbf{FMoW\cite{koh2021wilds,christie2018functional}}: Figure \ref{fig:fmow_data}. The images differ in region but also in resolution and scale. The distribution shift between FMoW and the pretraining data is large, therefore FmoW represents the ability of ERM++ to perform on non web-scraped data (see Section 5 of the main paper).

\smallbreak
\noindent\textbf{PCAM\cite{koh2021wilds,bandi2018detection}}: Figure \ref{fig:pcam_data}. The images are difficult to parse for an untrained human, but without tumors the images seems to have smaller and more dense cell structure. We also use this to test the generalization of ERM++ to perform on non-webscraped data (see Section 5 of the main paper) Images from Figure in \cite{koh2021wilds}.

\subsection{Attention Tuning Visualization}

We visualize attention tuning attention maps, and compare them to ERM++ w/out attention tuning and a pre-trained DINOv2 model in Figure \ref{fig:attn_tune}. We find attention tuning can pick up discriminative, but occluded, features in samples where ERM++ w/out attention tuning.

\section{Runtime Comparisons}

As discussed in the main paper Section 4.4; ERM++ achieves higher predictive performance than competing methods MIRO~\cite{cha2022domain} and DIWA~\cite{rame2022diverse} despite lower computational cost for training. The reason is reduced cost of hyper-parameter search; we use fixed hyper-parameters, borrowed from the  DomainBed framework, (see Section \ref{subsec:training_details} for more details ) while DIWA averages 20-60 models and MIRO search for \textit{4} $\lambda$ weight regularization values in each experiment. Assuming the worst case scenario of training two full passes (one on validation data for the training step cap in \textit{Auto-lr},  and one on full training data with validation data folded in \textit{Full Data}), and the same number of training steps as MIRO; ERM++ costs $\frac{1}{2}$ that of MIRO while obtaining better performance. In particular, this configuration represents Experiment 8 in Table 5 of the main paper. 

For each forward step MIRO there is an additional forward pass of the data through the model which is absent in ERM++. On the other hand, ERM++ does take a forward pass through the running average model to update batch normalization statistics, which is not done in former methods. This means that each forward pass is compute-equivalent for ERM++ and MIRO, for a given architecture.    

\section{Reproducibility}
\label{section:repro}

We provide code in a zip file along with this supplementary, and will open-source the code upon acceptance.

\subsection{Infrastructure}

We train on a heterogeneous cluster, primarily on NVIDIA A6000 GPU's. Each experiment is conducted on a single GPU with 4 CPUs. A single run could range from 12-48 hours, depending on number of steps trained. 

\subsection{Training details}
\label{subsec:training_details}

We follow the DomainBed~\cite{gulrajani2020search}  training procedure and add additional components from ERM++. 
In particular, we use the default hyper-parameters from DomainBed \cite{gulrajani2020search}, \eg, a batch size of 32 (per-domain), a learning rate of 5e-5, a ResNet dropout value of 0, and a weight decay of 0. We use the ADAM optimizer~\cite{kingma2014adam} optimizer with $\beta$ and $\epsilon$ values set default values from Pytorch 1.12. We extend the training cap to 4x the initial learning when \textbf{\textit{Auto-lr}} is used.  We train on all source domains except for one, validate the model on held-out data from the sources every 300 steps(20\% of the source data), and evaluate on the held-out domain. If using \textit{Full Data} we retrain using the full data. We use the same data augmentation techniques as ERM~\cite{gulrajani2020search}. %

\smallbreak
\noindent\textbf{ViT Training Details:} We follow a similar recipe for ViTs, with a few changes. First, we don't extend the training step cap by 4x on account of ViT's being over-parameterized and easy to overfit, relative to ResNet-50. Second, we use the LARS optimizer, which adjusts the learning rate according to the weight norm per-layer, and with a larger 1e-1 learning rate for the linear classifier during warmup ($1e-1$). The LARS optimizer decreases weight updates to stabilize training, and ViTs are unstable to train. 

\smallbreak
\noindent\textbf{Model Parameter Averaging details: }If we use Model Parameter Averaging(~\textit{MPA}), we begin to keep a running average at the 100th step. If we additionally use warm-start, we only optimize the classification head for the first 500 steps (2500 for ViT), and start \textit{MPA} 100 steps after that. For the Specialist Model Parameter Averaging(\textit{SMPA}) experiments (Table 9 of main paper), we first train a generalist model for 15000 steps , then train an independent model for each domain for another 1500 steps. At the end, we average parameters and re-compute batch norm running statistics. This recomputing of BN stats makes sure the averaged model has accurately computed batch norm statistics which may not be a simple average of  experts, due to the non-linearity of neural nets. 

\smallbreak
\noindent\textbf{Batch Normalization details:} With unfrozen batch normalization(~\textit{UBN}), we update the evaluation model BN statistics by averaging the model iterates first (from ~\textit{MPA}), then then forward propagating the current batch at each step through the evaluation model.  In this way, the BN running statistics and model used for inference match.

\noindent\textbf{Pseudocode for auto-lr:}

In Figure \ref{fig:learning_rate_decay}, we show pseudo-code for `Auto-lr' ERM++ component.  The learning rate is decreased by a factor
of 0.1 when the validation loss does not decrease. Training is stopped after the third consecutive
non-decreasing validation loss

\noindent\textbf{Sources of pre-trained weights:} We use torchvision 0.13.1 for vanilla ResNet-50 initialization. For augmix and ResNet-A1 initialized weights, we leverage  TIMM~\cite{rw2019timm} \footnote{ Augmix Weights :\url{https://github.com/rwightman/pytorch-image-models/releases/download/v0.1-weights/resnet50_ram-a26f946b.pth}} \footnote{ ResNet-A1 Weights :\url{https://github.com/rwightman/pytorch-image-models/releases/download/v0.1-rsb-weights/resnet50_a1_0-14fe96d1.pth}} .  

\noindent\textbf{A note on hyper-parameter search:} In this work, we focus on methodological improvements that do not depend on expensive hyper-parameter tuning, and as a result we use default learning rate, weight decay, etc. We demonstrate state-of-the-art performance  despite this, and greatly reduce the computational cost of training as a result. However, we believe there is substantial headroom for  improvement with further hyper-parameter tuning. 

\noindent\textbf{MIRO Implementation:} We directly follow the MIRO implementation and borrow the lambda weights values from~\cite{cha2022domain} when we combine MIRO with ERM++ in Table 2 of the main paper. ERM++ substantially improves the performance of MIRO.

\noindent\textbf{DIWA Implementation:}  We follow a simplified version of the DIWA~\cite{rame2022diverse}  algorithm due to computational reasons; we average the parameters of the three seeds of ERM++, with shared initialization of the linear classifier. The authors of DIWA show that about half of the performance boost comes from the first few models averaged (Figure 4 of ~\cite{rame2022diverse}), therefore this is a reasonable approximation of the method.

\noindent\textbf{SWAD Implementation}: We directly follow the SWAD implementation and hyper-parameters from \cite{cha2021swad}.

%% file: 9_tables_supp_no_se.tex
\begin{table*}[t]
\begin{center}

\begin{tabular}{l|cccc|l} \toprule
                  & art                       & clipart                   & product                   & real                      & avg                       \\ \midrule
ERM~\cite{gulrajani2020search}               & 63.1 & 51.9 & 77.2 & 78.1 & 67.6 \\
ERM + SWAD~\cite{cha2021swad}        & 66.1 & 57.7 & 78.4 & 80.2 & 70.6 \\
DIWA~\cite{rame2022diverse}             & 69.2                      & 59                        & 81.7                      & 82.2                      & 72.8                      \\
ERM + MIRO + SWAD~\cite{cha2022domain} & -                         & -                         & -                         & -                         & 72.4                      \\
ERM++             & 70.7 & \textbf{62.2} & 81.8 & 84.0 & 74.7 \\ \
ERM++ + MIRO           & \textbf{74.0} & 61.5 & \textbf{83.8} & \textbf{85.7} & \textbf{76.3} \\ \bottomrule
\end{tabular}
\end{center}
\caption{\textbf{OfficeHome:} Per-domain top-1 accuracy against reported results of recent top-performing methods SWAD, DIWA, and MIRO. \cite{cha2022domain} does not report per-domain performance for MIRO, so we only show average for that case. DIWA doesn't report standard errors. ERM++ not only greatly increases performance relative to SWAD, DIWA, and MIRO but also reduce variance between runs. The largest gains are on the held-out domain with the largest domain shift(clipart), illustrating the ability of ERM++ to improve performance on difficult DG tasks.}
\label{tab:oh}
\end{table*}

\begin{figure*}[t]
\begin{center}
\includegraphics[width=0.81\linewidth]{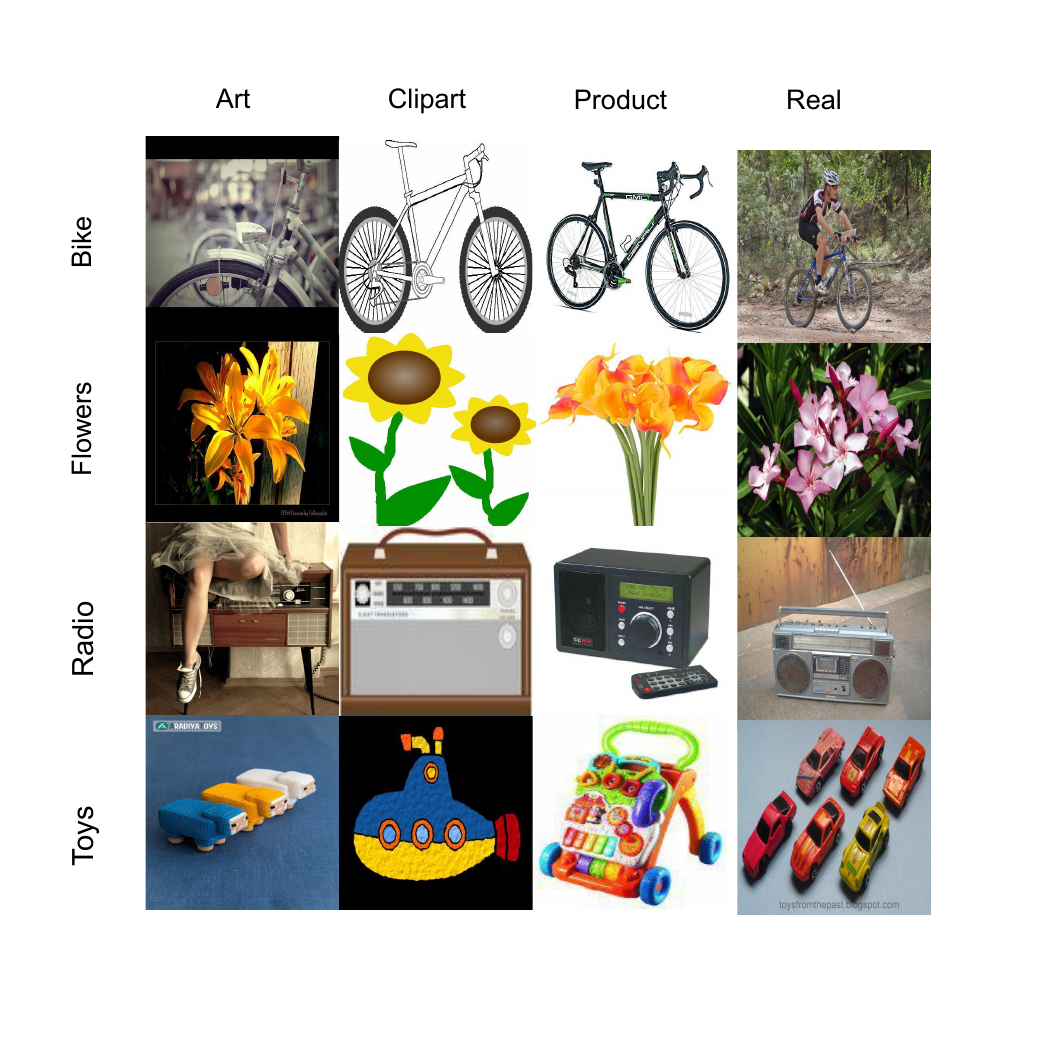}
\caption{\textbf{OfficeHome}: Samples from the OfficeHome\cite{venkateswara2017deep} dataset, from each domain and selected classes. The dataset focuses on household objects. The domain shifts are in low-level style mostly, and there is little spatial bias. }
\label{fig:oh_data}
\end{center}
\end{figure*}

\begin{table*}[t]
\centering
\setlength{\tabcolsep}{2pt}

\begin{tabular}{l|cccccc|c} \toprule
                  & painting                  & clipart                   & info                      & real                      & quickdraw                 & sketch                    & avg                       \\ \midrule
ERM~\cite{gulrajani2020search}               & 50.1 & 63.0 & 21.2 & 63.7 & 13.9 & 52.9 & 44.0 \\
ERM + SWAD~\cite{cha2021swad}        & 53.5 & 66.0 & 22.4 & 65.8 & 16.1 & 55.5 & 46.5 \\
DIWA~\cite{rame2022diverse}              & 55.4                      & 66.2                      & 23.3                      & 68.7                      & 16.5                      & 56                        & 47.7                      \\
ERM + MIRO + SWAD~\cite{cha2022domain} & -                         & -                         & -                         & -                         & -                         & -                         & 47.0 \\
ERM++             & 58.4 & \textbf{71.5} & 26.2 & 70.7 & \textbf{17.3} & \textbf{60.5} & \textbf{50.8} \\ 
ERM++ + MIRO &           \textbf{58.5} & 71.0 & \textbf{26.5} & \textbf{71.1} & 15.9 & 59.5 & 50.4\\ \bottomrule
\end{tabular}
\caption{\textbf{DomainNet:} Per-domain top-1 accuracy against reported results of recent top-performing methods SWAD, DIWA, and MIRO. \cite{cha2022domain} does not per-domain performance for MIRO, so we only show average for that case. DIWA doesn't report standard errors. ERM++ not only greatly increases performance relative to SWAD, DIWA, and MIRO but also reduce variance between runs. Similar to results on OfficeHome (Table \ref{tab:oh}), the largest performance gains(of larger than 4\%) are on domains very different from the source domain(clipart and sketch). This suggests ERM++ is less sensitive to texture bias than ERM~\cite{gulrajani2020search}. The bias of MIRO to the pre-trained weights manifests in slightly higher performance on close to ImageNet domains like real when combined with ERM++, at the slight expense of performance on other domains. }
\label{tab:dn}
\end{table*}

\begin{figure}[h!]
\centering
\includegraphics[width=\linewidth]{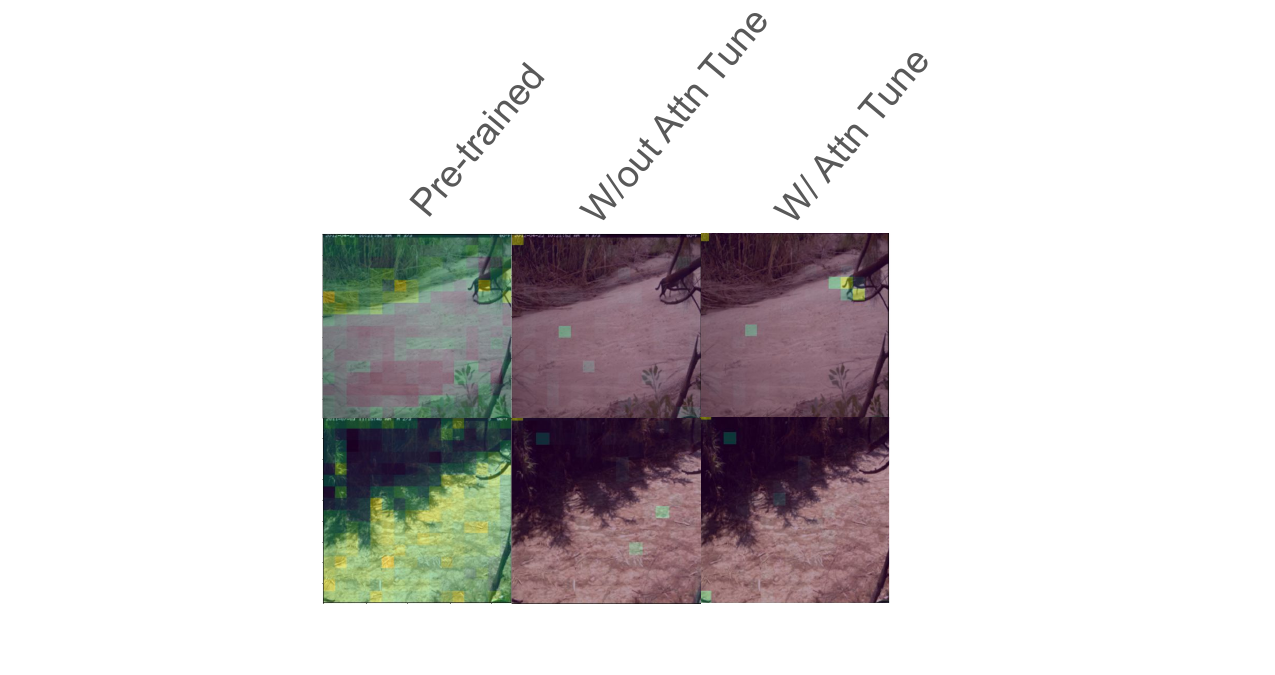}
\caption{ Examples of Attention Tuning Visualization of DINOv2 model. We average over all attention heads in the final attention block.  On a pretrained model, attention is scattered. On both an attention tuned and full fine-tuned model, attention is more focused than with a pre-trained model. However, on some samples (representive samples pictured here) full fine-tuning misses discriminative but occluded animal features. On the top-right images, the attention tuning picks up the dog. In the bottom-right, the attention-tuned model picks up a tail in the lower-left corner.   }
\label{fig:attn_tune}
\end{figure}

\begin{figure*}[t]
\begin{center}
\includegraphics[width=0.93\linewidth]{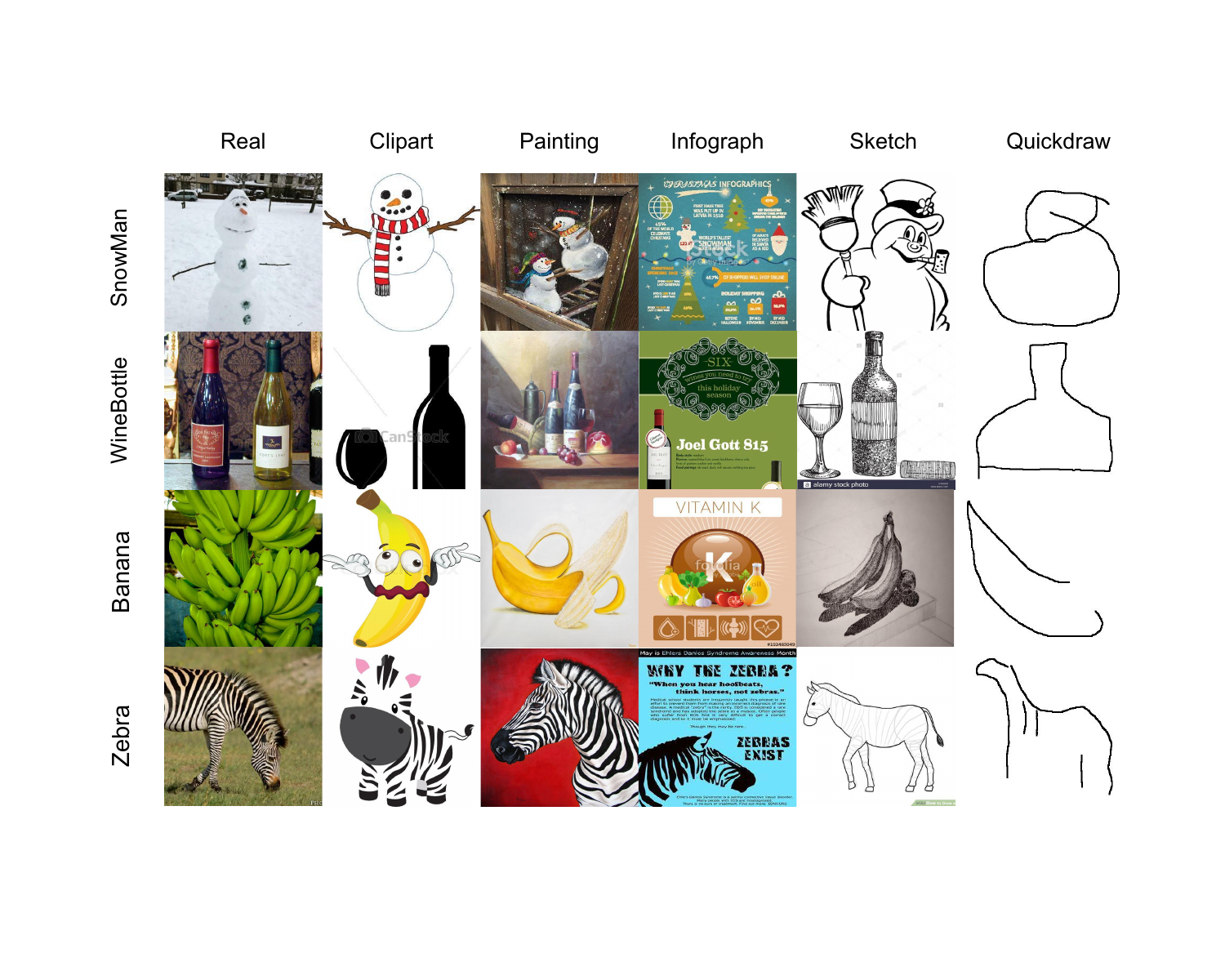}
\end{center}
\caption{\textbf{DomainNet:} Samples from the DomainNet\cite{peng2019moment} dataset. While the real domain is quite similar to what one might expect in ImageNet, the distribution shifts are quite substantial in other domains. Quickdraw and Infograph are particularly challenging, so the 1-3\% gains of ERM++ on these domains is meaningful (Table \ref{tab:dn}). While most domains contain primarily shifts in low level statistics (for example, real to painting), Infograph also has many non-centered objects. }
\label{fig:dn_data}
\end{figure*}

\begin{table*}[t]
\begin{center}

\begin{tabular}{l|cccc|c} \toprule
                  & caltech101                & labelme                   & sun09                     & voc2007                   & avg                       \\ \midrule
ERM~\cite{gulrajani2020search}               & 97.7 & \textbf{64.3} & 73.4 & 74.6 & 77.3 \\
ERM + SWAD~\cite{cha2021swad}       & 98.8 & 63.3 & \textbf{75.3} & \textbf{79.2} & 79.1 \\
DIWA~\cite{rame2022diverse}             & \textbf{98.9}                      & 62.4                      & 73.9                      & 78.9                      & 78.6                      \\
ERM + MIRO + SWAD~\cite{cha2021swad} & -                         & -                         & -                         & -                         & \textbf{79.6} \\
ERM++             & 98.7 & 63.2 & 71.6 & 78.7 & 78.0 \\ 
ERM++ + MIRO           & 99.0 & 62.4 & 71.8 & 78.3 & 77.9 \\ \bottomrule

\end{tabular}
\caption{\textbf{VLCS:} Per-domain top-1 accuracy against reported results of recent top-performing methods SWAD, DIWA, and MIRO. \cite{cha2022domain} does not per-domain performance for MIRO, so we only show average for that case. DIWA doesn't report standard errors. Although overall performance on VLCS is lower than competing methods, we can see that this drop primarily comes from lower performance on sun09.  Furthermore, there are many ambiguous images in the LabelMe domain (see Figure \ref{fig:broken_vlcs}), raising questions about the usefulness of trying to train on this domain.  }
\label{tab:vlcs}
\end{center}
\end{table*}

\begin{figure*}[t]
\begin{center}
\includegraphics[width=0.75\linewidth]{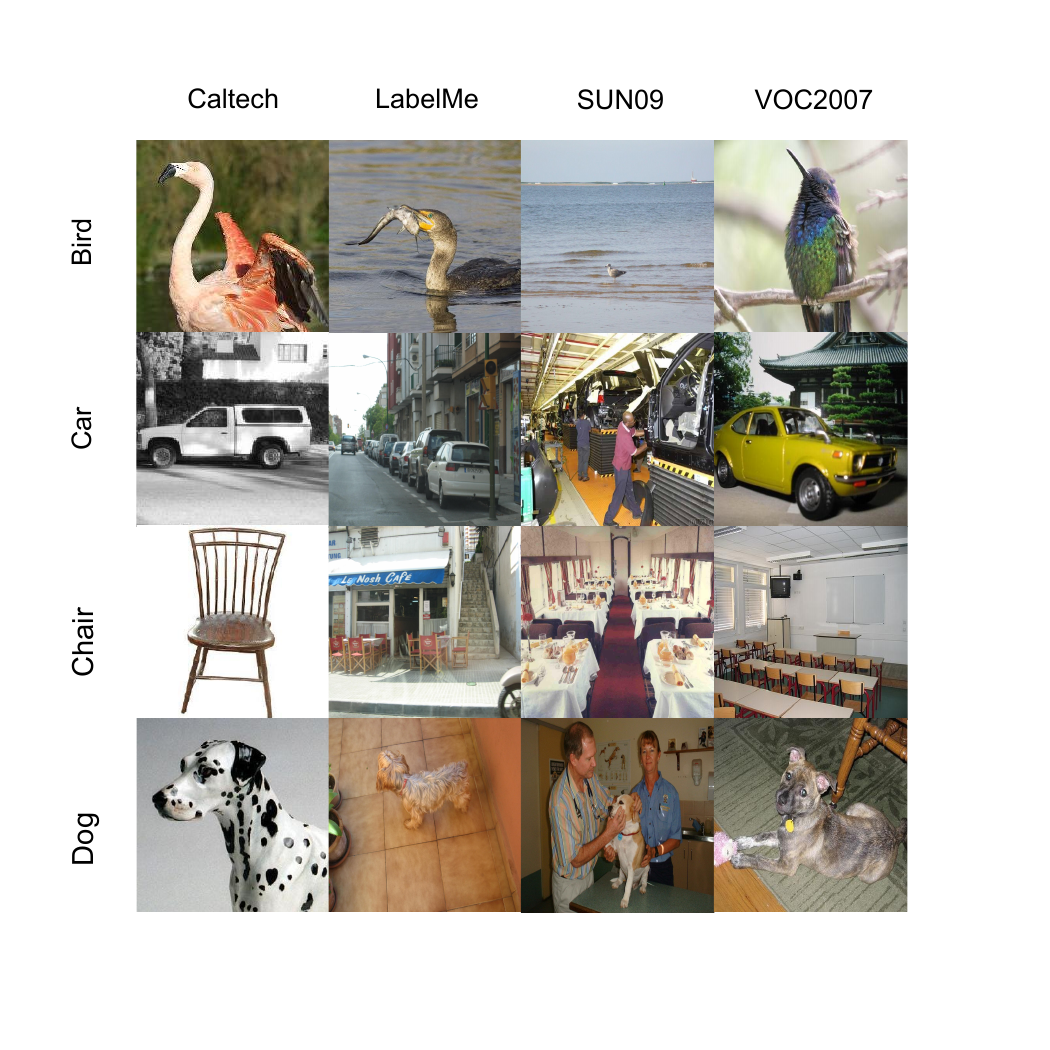}
\caption{\textbf{VLCS:} Sample from the VLCS dataset\cite{fang2013unbiased} The low-level statistics are quite similar between domains, however spatial biases differ between domains. Caltetch objects are quite centered, while other domains do not have this trait. For example the LabelMe domain has cars along the side of the image, and there are many chairs in the VOC2007 domain. Furthermore, in some cases the size of the objects differs dramatically.  Finally, there are many ambiguous images in the LabelMe domain (see Figure \ref{fig:broken_vlcs}), raising questions about the usefulness of trying to train on this domain.  }
\label{fig:vlcs_data}
\end{center}
\end{figure*}

\begin{table*}[t]
\begin{center}
\setlength{\tabcolsep}{2pt}

\begin{tabular}{l|cccc|c} \toprule
                  & Loc. 100              & Loc. 38               & Loc. 43               & Loc. 46               & Average                   \\ \midrule
ERM~\cite{gulrajani2020search}               & 54.3 & 42.5 & 55.6 & 38.8 & 47.8 \\
ERM + SWAD~\cite{cha2021swad}       & 55.4 & 44.9 & 59.7 & 39.9 & 50.0 \\
DIWA~\cite{rame2022diverse}             & \textbf{57.2}                      & 50.1                      & 60.3                      & 39.8                      & 51.9                      \\
ERM + MIRO + SWAD~\cite{cha2022domain} & -                         & -                         & -                         & -                         &               52.9          \\
ERM++             & 48.3 & \textbf{50.7} & \textbf{61.8} & \textbf{43.9} & 51.2 \\
ERM++ + MIRO           & \textbf{60.81} & 48.8 & 61.1 & 42.7 & \textbf{53.4} \\ \bottomrule
\end{tabular}
\caption{\textbf{TerraIncognita:} Per-domain top-1 accuracy against reported results of recent top-performing methods SWAD, DIWA, and MIRO. \cite{cha2022domain} does not per-domain performance for MIRO, so we only show average for that case. DIWA doesn't report standard errors. ERM++ outperforms other methods on 3 out of 4 held out domains despite slighly underperforming on average. However, we point out that ERM++ w/MIRO  outperforms both DIWA and MIRO, and improves ERM++ by a further 2\%.    }
\label{tab:ti}
\end{center}

\end{table*}

\begin{figure*}[t]
\begin{center}
\includegraphics[width=0.83\linewidth]{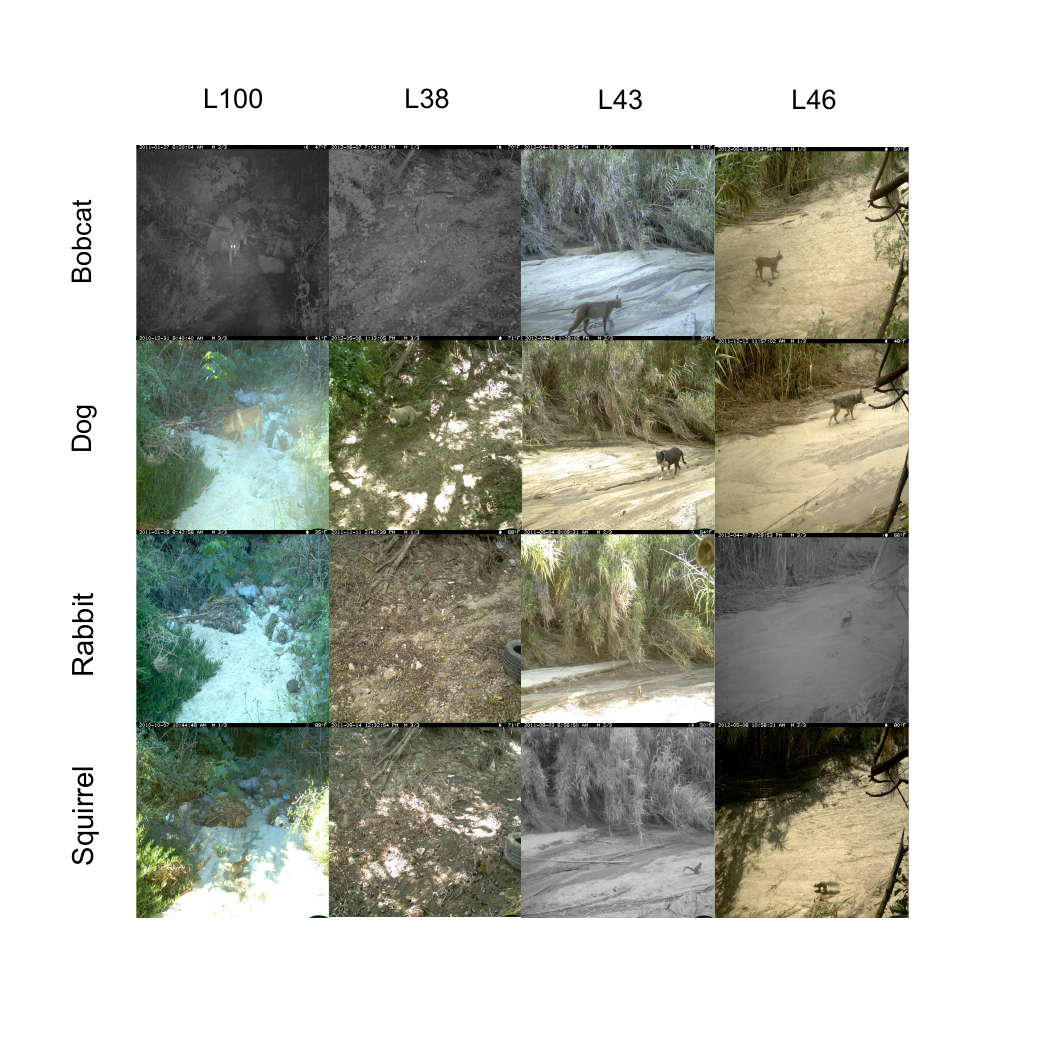}
\caption{\textbf{TerraIncognita}: Samples from the TerraIncognita\cite{beery2018recognition} dataset, from each domain and selected classes. The background stays consistent, and the animal object frequently takes up a small portion of the frame. At night the images are black-and-white. This dataset matches realistic deployment scenarios well.  }
\label{fig:ti_data}
\end{center}
\end{figure*}

\begin{table*}[t]
\begin{center}

\begin{tabular}{l|cccc|c} \toprule
                  & art\_painting              & cartoon                   & photo                     & sketch                    & avg                       \\ \midrule
ERM~\cite{gulrajani2020search}              & 84.7 & 80.8 & 97.2 & 79.3 & 84.2 \\
ERM + SWAD~\cite{cha2021swad}        & 89.3 & 83.4 & 97.3 & 82.5 & 88.1 \\
DIWA~\cite{rame2022diverse}              & 90.6                      & 83.4                      & 98.2                      & 83.8                      & 89                        \\
ERM + MIRO + SWAD~\cite{cha2022domain} & -                         & -                         & -                         & -                         & 88.4 \\
ERM++             & \textbf{90.6} & 83.7 & 98.1 & \textbf{86.6} & \textbf{89.8} \\ 
ERM++ + MIRO           & 90.2 & \textbf{83.8} & \textbf{98.6} & 82.4 & 88.8 \\ \bottomrule

\end{tabular}
\caption{\textbf{PACS:} Per-domain top-1 accuracy against reported results of recent top-performing methods SWAD, DIWA, and MIRO. \cite{cha2022domain} does not per-domain performance for MIRO, so we only show average for that case. DIWA doesn't report standard errors. ERM++ leads to substantial improvement over prior work. As in other dataset (OfficeHome, DomainNet), large performance gains are made on the sketch domain.  }
\label{tab:pacs}
\end{center}

\end{table*}

\begin{figure*}[t]
\begin{center}
\includegraphics[width=0.81\linewidth]{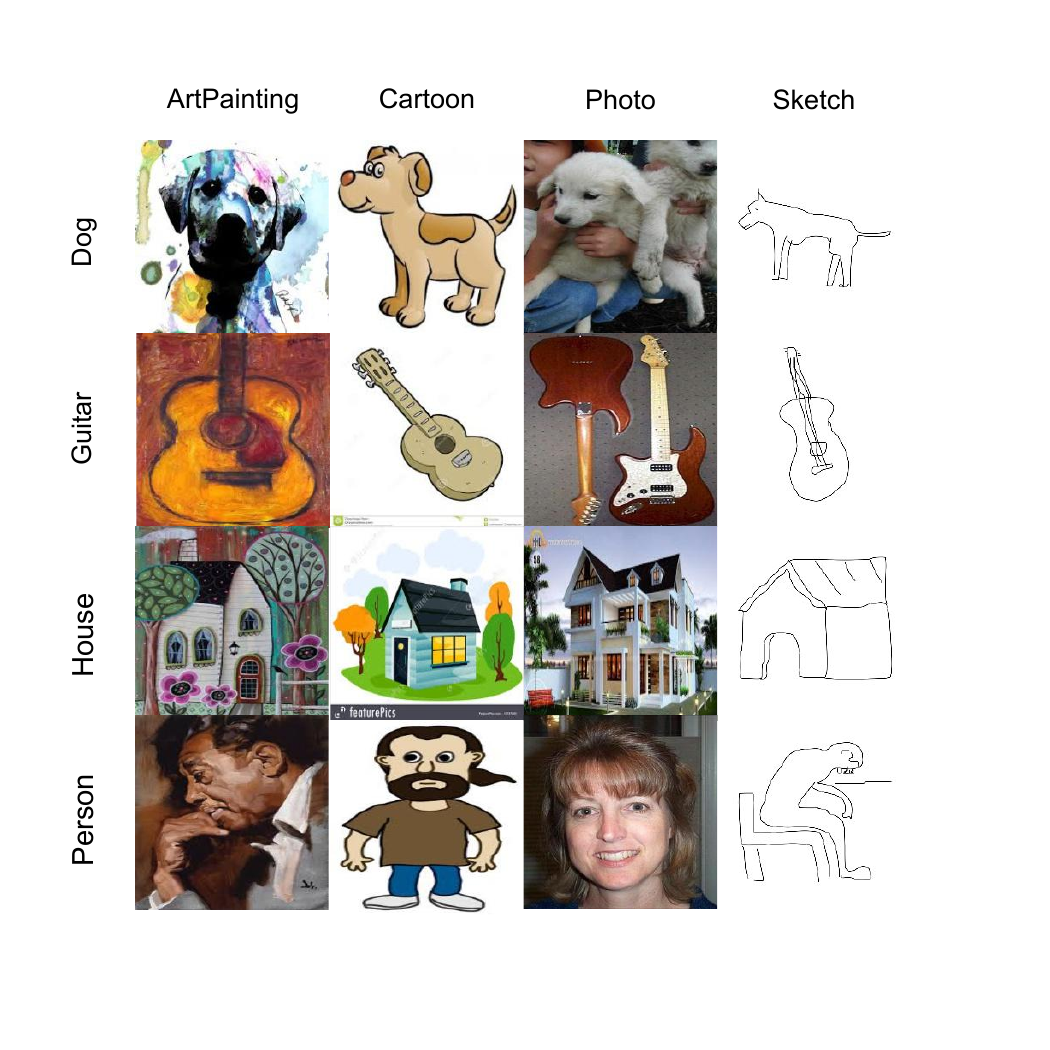}
\caption{\textbf{PACS:} Samples from the PACS dataset \cite{li2017deeper}, from each domain and selected classes. The subjects tend to be centered, and the sketches are more realistic than the quickdraw setting in DomainNet. Though the domians are similar to that of DomainNet, PACS has fewer than 10000 samples compared to 586000 of DomainNet. Therefore PACS tests the capabilities of ERM++ on smaller data.}
\label{fig:pacs_data}
\end{center}
\end{figure*}

\begin{figure*}[t]
\begin{center}
\includegraphics[width=0.8\linewidth]{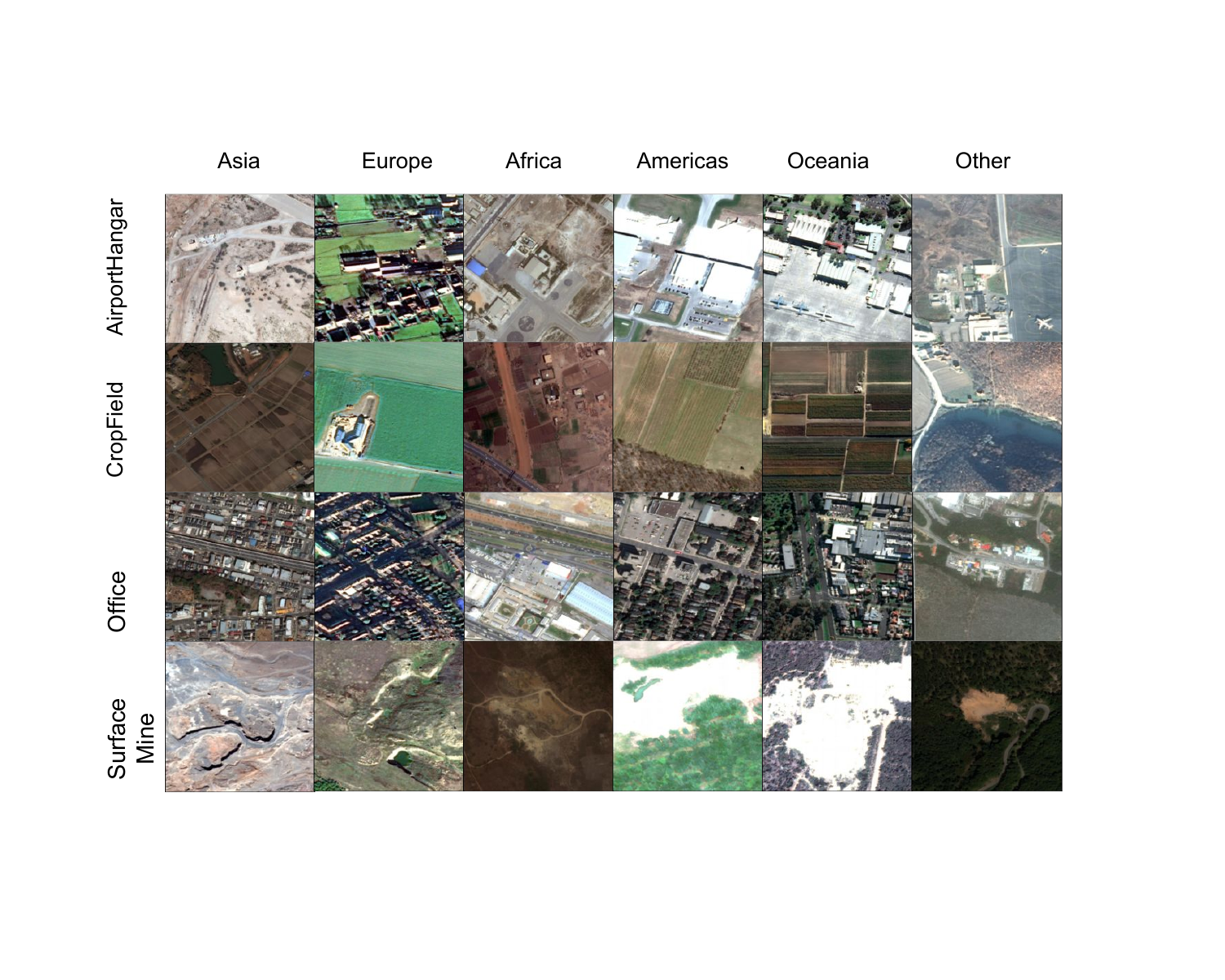}
\caption{\textbf{FMoW:} Samples from the FMoW\cite{christie2018functional,koh2021wilds} dataset, from each domain and selected classes. The images differ in region but also in resolution and scale. The distribution shift between FMoW and the pretraining data is large, therefore FmoW represents the ability of ERM++ to perform on non web-scraped data (see Section 5.4 of the main paper). }
\label{fig:fmow_data}
\end{center}
\end{figure*}

\begin{figure*}[t]
\begin{center}
\includegraphics[width=0.8\linewidth]{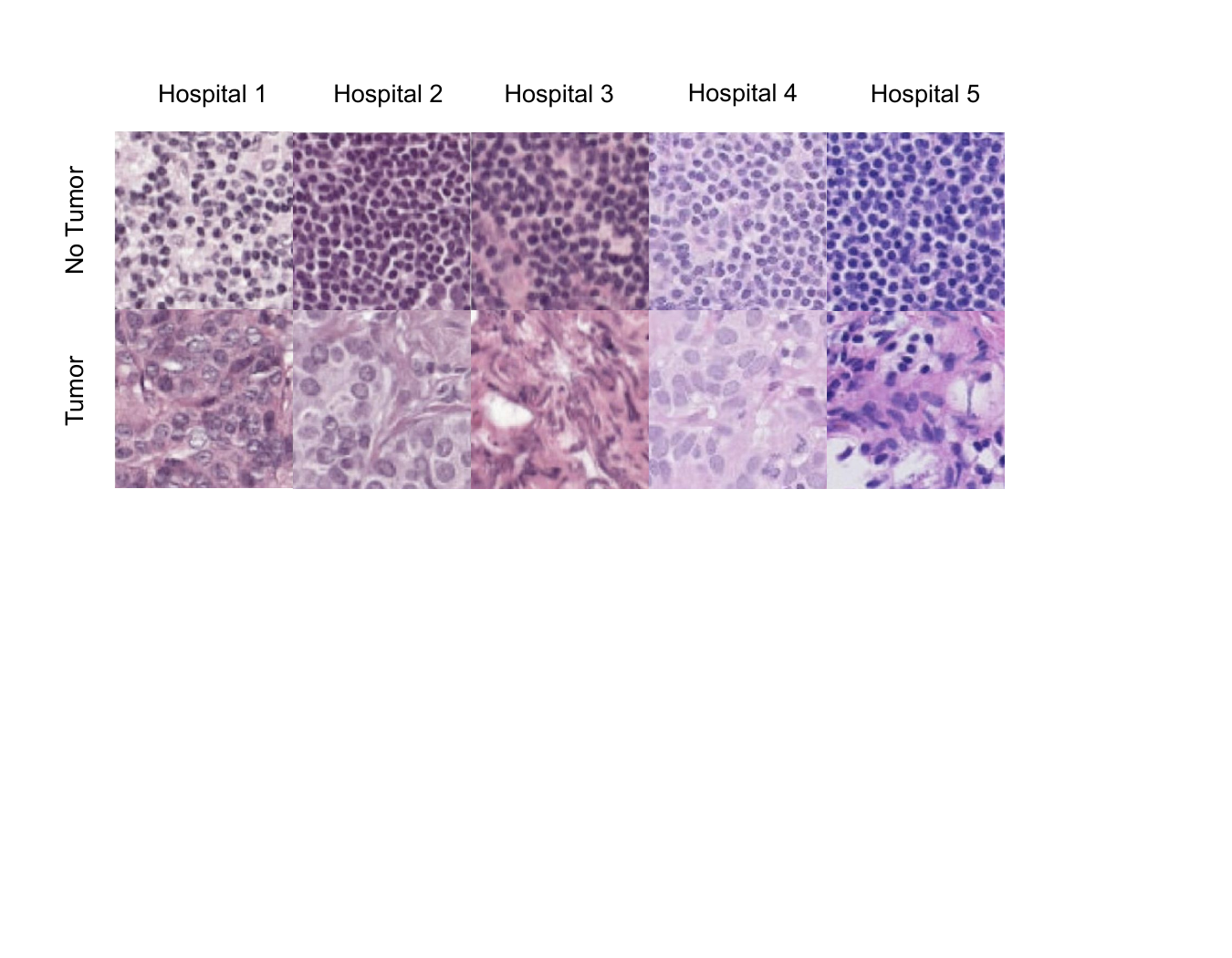}
\vspace{-70pt}
\caption{\textbf{PCAM:} Samples from the PatchCamelyon\cite{bandi2018detection,koh2021wilds} dataset, from each domain and both classes. The images are difficult to parse for an untrained human, but without tumors the images seems to have smaller and more dense cell structure. Images from \cite{koh2021wilds} paper figure.}
\label{fig:pcam_data}
\end{center}
\end{figure*}

\begin{figure*}[t]
\begin{center}
\includegraphics[width=0.8\linewidth]{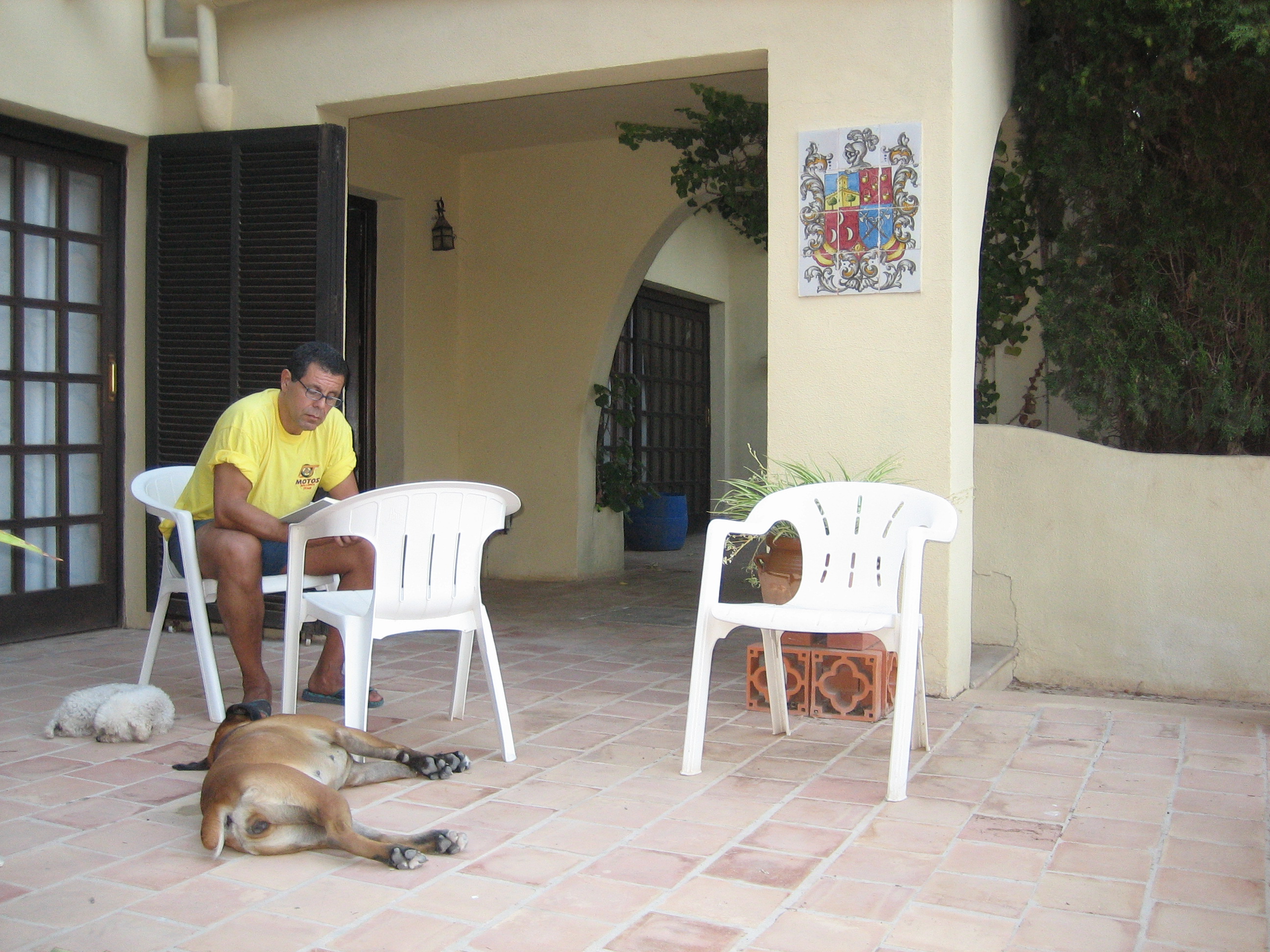}
\caption{\textbf{Sample from LabelMe Domain in VLCS:} Is this a dog, person, or chair? Many samples in the LabelMe domain of VLCS are ambiguous but assigned a label (in this case, dog). This raises questions about the usefulness of training on this domain.  }
\label{fig:broken_vlcs}
\end{center}
\end{figure*}